\documentclass{article}

\usepackage[numbers]{natbib}
\usepackage[preprint]{llamagen}

\usepackage[dvipsnames]{xcolor}         
\definecolor{linkColor}{rgb}{0.18,0.39,0.62}
\usepackage[utf8]{inputenc} 
\usepackage[T1]{fontenc}    
\usepackage[colorlinks=true,linkcolor=linkColor,citecolor=linkColor,filecolor=linkColor,urlcolor=linkColor]{hyperref}       
\usepackage{url}            
\usepackage{booktabs}       
\usepackage{amsfonts}       
\usepackage{nicefrac}       
\usepackage{microtype}      

\RequirePackage{algorithm}
\RequirePackage{algorithmic}

\bibpunct{[}{]}{;}{a}{}{,}

\usepackage{multirow}
\usepackage{amsmath}
\usepackage{capt-of}
\usepackage{tabularx}
\usepackage{epsfig}
\usepackage{amssymb}
\usepackage{amsfonts}
\usepackage{subcaption} 
\usepackage{booktabs}
\usepackage{scalerel}
\usepackage[inline]{enumitem}
\usepackage{listings}
\usepackage{varwidth}
\usepackage[export]{adjustbox}
\usepackage{tikz}
\usetikzlibrary{tikzmark}

\usepackage{stmaryrd}
\usepackage{bbm}
\usepackage{wrapfig}
\usepackage{pifont}

\definecolor{deepblue}{rgb}{0,0,0.5}
\definecolor{officeblue}{RGB}{0,102,204}
\definecolor{deepred}{rgb}{0.6,0,0}
\definecolor{deepgreen}{rgb}{0,0.5,0}
\definecolor{mybrickred}{RGB}{182,50,28}

\definecolor{fillcolor}{RGB}{216,217,252}

\usepackage{pifont}

\newcommand\our{LlamaGen}

\title{Autoregressive Model Beats Diffusion: Llama for Scalable Image Generation}

\author{
  \vspace{-15mm}\\
  \textbf{Peize Sun$^{1}$
  \quad Yi Jiang$^{2\dag}$
  \quad Shoufa Chen$^{1}$
  \quad Shilong Zhang$^{1}$
  \quad Bingyue Peng$^2$} \vspace{2mm} \\
  \textbf{
  Ping Luo$^{1*}$
  \quad Zehuan Yuan$^{2}$\thanks{: Corresponding authors, $\dag$: project lead }}\vspace{2mm} \\
  $^1$The University of Hong Kong~~\quad\quad $^2$ByteDance \vspace{2mm} \\
  Codes and models:~\, \url{https://github.com/FoundationVision/LlamaGen}
  \vspace{-5mm} \\
}

\begin{document}
\maketitle

\begin{figure}[ht]
\vspace{-5mm}
\begin{center}
\includegraphics[width=1.00\linewidth]{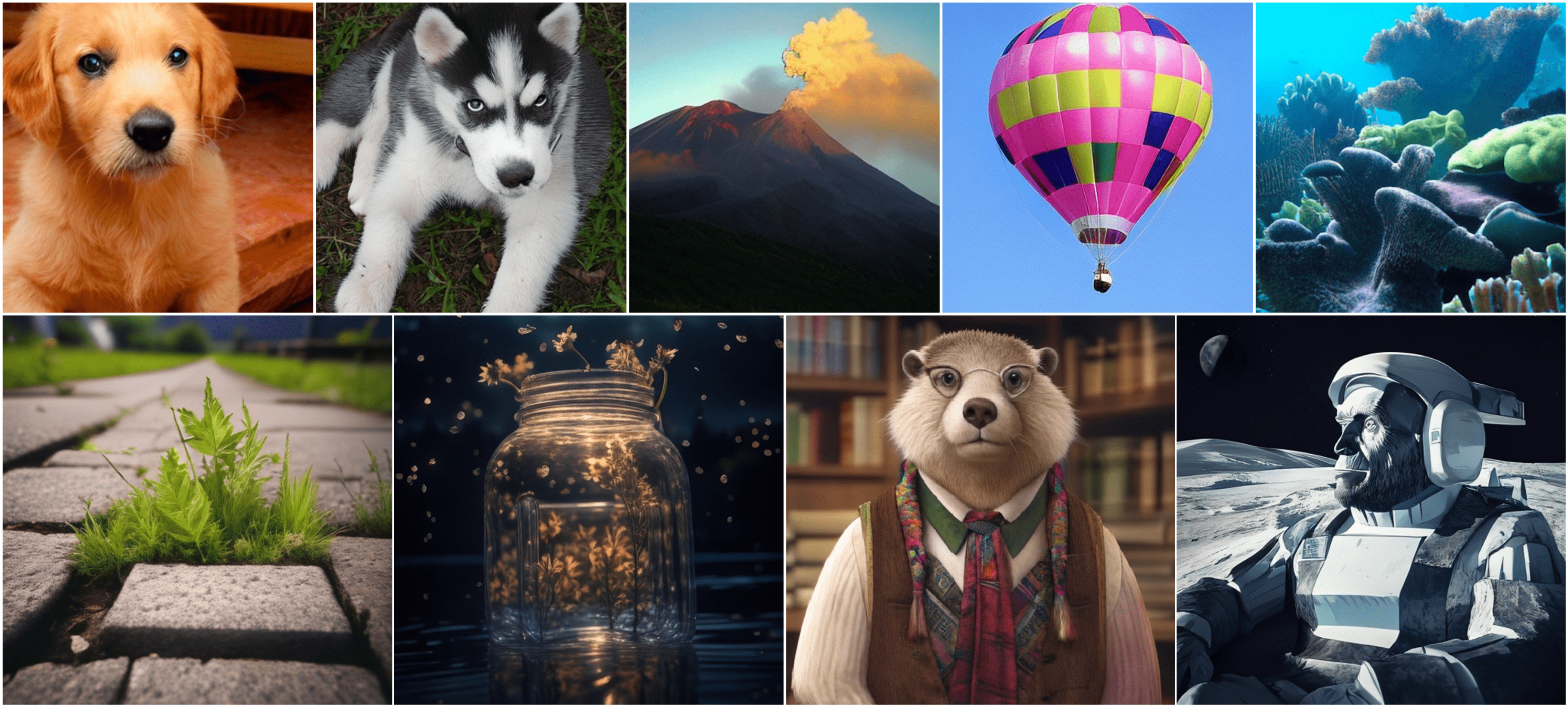}
\end{center}
\vspace{-2mm}
\caption{\textbf{Image generation with vanilla autoregressive models}. We show samples from our class-conditional image (top row) and text-conditional image (bottom row) generation models.}
\label{fig:abs}
\end{figure}

\begin{abstract}
We introduce \our, a new family of image generation models that apply original ``next-token prediction'' paradigm of large language models to visual generation domain. It is an affirmative answer to whether vanilla autoregressive models, e.g., Llama, without inductive biases on visual signals can achieve state-of-the-art image generation performance if scaling properly. We reexamine design spaces of image tokenizers, scalability properties of image generation models, and their training data quality. The outcome of this exploration consists of: (1) An image tokenizer with downsample ratio of 16, reconstruction quality of 0.94 rFID and codebook usage of 97\% on ImageNet benchmark. (2) A series of class-conditional image generation models ranging from 111M to 3.1B parameters, achieving 2.18 FID on ImageNet 256×256 benchmarks, outperforming the popular diffusion models such as LDM, DiT. (3) A text-conditional image generation model with 775M parameters, from two-stage training on LAION-COCO and high aesthetics quality images, demonstrating competitive performance of visual quality and text alignment. (4) We verify the effectiveness of LLM serving frameworks in optimizing the inference speed of image generation models and achieve 326\% - 414\% speedup. We release all models and codes to facilitate open-source community of visual generation and multimodal foundation models.
\end{abstract}

\section{Introduction}
\label{sec:intro}
Built upon autoregressive models, large language models (LLMs)~\citep{transformer,bert, gpt1,t5,gpt2,gpt3,opt} generate the text by predicting the next token in a sequence. This ``next-token prediction'' paradigm presents unprecedented capabilities in solving language tasks in a human-like conversational manner~\citep{gpt3.5,openai2022chatgpt,gpt4,google2023bard,anthropic2023claude,bloom,Llama1,Llama2,qwen,baichuan,internlm,deepseek} and incredible scalability~\citep{scalinglaw,scalingar,chinchilla,emergent,revisitingscalinglaws,palm,palm2}, demonstrating a promising path toward general-purpose artificial intelligence models.

Witnessed the scalability of autoregressive models on large language models, pioneering works attempt to explore autoregressive models in image generation, for example, VQVAE~\citep{vqvae, vqvae2}, VQGAN~\citep{vqgan, rq}, DALL-E~\citep{dalle1}, Parti~\citep{vit-vqgan, parti}. They introduce image tokenizers to convert continuous images to discrete tokens, and apply autoregressive models to generate image tokens in the way of next-token prediction.
They demonstrate strong performance among their contemporaries~\citep{biggan,ddpm,adm} in the year before 2022. However, their open-source communities are not well developed, which largely limits their further improvements.

At the same period, another image generation method, diffusion models~\citep{scorebased,ddpm,ddim,adm,glide,dpm-solver,cascade-diffusion,cfg,ldm,dalle2,imagen,ldm} develop rapidly. Along with their open-source communities, they dominate the field of visual generation up to today. 
However, diffusion models share distinct paradigms with autoregressive language models, which poses a huge challenge to building a unified model between language and vision.

In this work, we are committed to pushing the envelope of autoregressive models on image generation further: \textit{continuing its research methodology and contributing to open-source community.} Reviewing the literature on image generation in the year before 2024,  we identify three keys to existing advanced models~\citep{dit,sdxl,raphael,pixart,gentron,dalle3,playgroundv2.5,sd3}: 1) well-designed image compressors, 2) scalable image generation models and 3) high-quality training data. Motivated by this, we reexamine the designs of image tokenizers (image compressors for autoregressive models), the scalability properties of image generation models, and the effects of training data. 

Towards a potential unified model between
language and vision, our design is reducing the inductive biases on visual signals and adopting the same architecture as LLM. This belongs to a different research philosophy with recent works~\citep{maskgit,magvit2,var} that modify the architectures under the guidance of vision-oriented designs. For example, MaskGIT~\citep{maskgit}, MAGVIT~\citep{magvit,magvit2} adopt the masked image modeling strategy, VAR~\citep{var} uses hierarchical multi-scale property. Although they have succeeded in achieving leading image generation performance, and even better than diffusion models, it is still not clear whether the original language model architectures are capable of this. Instead, our work reveals that vanilla autoregressive models that apply the exactly same ``next-token prediction'' as language models are also able to achieve state-of-the-art image generation performance. As a bonus, we can leverage the techniques~\citep{flashattention,deepspeed,megatron,fsdp,vllm,speculative,bitsandbytes} developed in LLM community to optimize the training recipes and inference speeds of our models.

In summary, our contributions to the community include:

\begin{enumerate}[topsep=3.5pt,itemsep=3pt,leftmargin=20pt]
\item \textbf{Image tokenizer:} An image tokenizer with downsample ratio of 16, achieves reconstruction quality of 0.94 rFID and codebook usage of  97\% on ImageNet benchmark. With the downsample ratio of 8, our tokenizer is competitive or even better than continuous VAE~\citep{ldm,sdxl,consistencydecoder} used in diffusion models. This shows that discrete representation in image tokenizers is no longer the bottleneck of the image reconstruction.
\item \textbf{Scalable image generation model:} A series of class-conditional image generation models, ranging from 111M to 3.1B parameters, are developed based on Llama architecture~\citep{Llama1,Llama2}. The largest model realizes 2.18 FID on ImageNet 256$\times$256 benchmarks, outperforming the popular diffusion models such as LDM~\citep{ldm}, DiT~\citep{dit}. This shows that vanilla autoregressive models without inductive biases on visual signals can serve as the basis of image generation systems.
\item \textbf{Hiqh-quality training data:} A text-conditional image generation model with 775M parameters, is firstly trained on a 50M subset of LAION-COCO~\citep{laion_coco} and then fine-tuned on 10M internal high aesthetics quality images. It demonstrates competitive performance of visual quality and text alignment. 
\item \textbf{Optimized inference speed:} We adopt vLLM~\citep{vllm}, one of the most popular LLM serving frameworks, to optimize the inference speed of our image generation models, and remarkable 326\% - 414\% speedup is achieved.
\end{enumerate}

We release all models and codes to facilitate the open-source community of visual generation and multimodal foundation models. It is worth noticing that our released models are still behind state-of-the-art visual generation models based on diffusion models~\citep{large-dit,sd3,sora}. When more training data and computation resources are available in the future, large-scale AR-based visual generation models, e.g., above 7B parameters, will be explored.

\section{Autoregressive Models for Image Generation} 
\subsection{Overview}
Firstly, image pixels $x\in\mathbb{R}^{H\times W\times3}$ are quantized into $q\in\mathbb{Q}^{h\times w}$ discrete tokens by the image tokenizer~\citep{vqvae,vqgan,vit-vqgan}, where $h$$=$$H/p$, $w$$=$$W/p$, $p$ is downsample ratio of the image tokenizer, $q^{(i,j)}$ is indices of the image codebook. Then, these image tokens are reshaped to a sequence of $h$$\cdot$$w$ tokens in raster scan ordering and used to train Transformer~\citep{transformer}-based autoregressive models.

During image generation, image tokens $(q_1, q_2, \dots, q_{h\cdot w})$ are generated by autoregressive models~\citep{gpt1,gpt2,gpt3,Llama1} in the way of next-token prediction $\prod_{t=1}^{h\cdot w} p(q_t \mid q_{<t},c)$, where $c$ is class label embedding or text embedding. Finally, these image tokens are converted to image pixels by the image tokenizer decoder.

\subsection{Image Tokenizer}
\paragraph{Quantized-Autoencoder architecture.}
We use the same architecture as VQGAN~\citep{vqgan}, encoder-quantizer-decoder. The encoder and the decoder are ConvNet with downsample ratio $p$. The quantizer contains a codebook $Z \in \mathbb{R}^{K\times C}$ with $K$ learnable vectors.
The encoder projects image pixels $x$ to the feature map $f$. The quantization process maps each vector $f^{(i,j)}$ in the feature map to the code index $q^{(i,j)}$ of its nearest vector $z^{(i,j)}$ in the codebook.
During decoding, the code index $q^{(i,j)}$ is remapped to the feature vector $z^{(i,j)}$ and the decoder converts these feature vectors back to the image pixels $\hat{x}$.

The codebook has critical effects on image tokenization performance. Following~\citep{vit-vqgan},  we use ${\ell}_2$-normalization to codebook vectors, low codebook vector dimension $C$, and large codebook size $K$. These designs significantly improve reconstruction quality and codebook usage. More details will be discussed in experiments.

\paragraph{Training losses.} 

Since quantization is a non-differentiable operation, a straight-through gradient estimator~\citep{estimator} is used to preserve the gradient from the decoder to the encoder $z = \text{sg}[z - f] + f$, $\text{sg}[\cdot]$ is stop-gradient operation. For codebook learning, $\mathcal{L}_{\text{VQ}} = \|\text{sg}[f] - z\|_2^2 + \beta\|f - \text{sg}[z]\|_2^2$, where the second term is commitment loss~\citep{vqvae} to force feature vectors extracted from the encoder to be close to codebook vectors, $\beta$ is commitment loss weight. For simplicity, we don't add entropy loss~\citep{magvit,maskgit} in codebook learning.

For image reconstruction training, $\mathcal{L}_{\text{AE}} = {\ell}_2(x, \hat{x}) + \mathcal{L}_{\text{P}}(x, \hat{x}) + \lambda_{\text{G}} \mathcal{L}_{\text{G}}(\hat{x})$, 
where ${\ell}_2$ is a reconstruction loss on image pixels, $\mathcal{L}_{\text{P}}(\cdot)$ is a perceptual loss from LPIPS~\citep{lpips}, $\mathcal{L}_{\text{G}}(\cdot)$ is an adversarial loss from a PatchGAN~\citep{pix2pix} discriminator trained at the same time with the image tokenizer, and $\lambda_{\text{G}}$ is adversarial loss weight.

\subsection{Image Generation by Autoregressive Models}
\paragraph{Llama architecture.} Our model architecture is largely based on Llama~\citep{Llama1,Llama2}, applying pre-normalization using RMSNorm~\citep{rmsnorm}, SwiGLU activation function~\citep{swiglu}, and rotary positional embeddings~\citep{rope}. Specifically, we use 2D RoPE in at each layer of our model, following the implementation of~\citep{unified-io2,eva2}. We do not use the technique of AdaLN~\citep{dit} to keep our structure the same as LLM.

\paragraph{Class-conditional image generation.} The class embedding is indexed from a set of learnable embeddings~\citep{dit,vqgan} and is used as the prefilling token embedding. Starting from this token embedding, the model generates the sequence of image tokens by next-token prediction way, and stops at the location of the pre-defined maximum length.

\paragraph{Text-conditional image generation.} To integrate the text condition into autoregressive models, we use FLAN-T5 XL~\citep{flan_t5} as the text encoder, the encoded text feature is projected by an additional MLP~\citep{pixart,gentron} and is used as prefilling token embedding in autoregressive models. We note that this design is not an ultimate design for multimodal foundation models, where a unified vocabulary is established between language and vision~\citep{unified-io2,gemini}. We leave it for future research.

\paragraph{Classifier-free guidance.} Developed in the diffusion model community, classifier-free guidance~\citep{cfg} is well-known for its improving visual quality and text-image alignment. We adopt it in our models. During training, the conditional is randomly dropped and is replaced by a null unconditional embedding~\citep{dit, pixart}. In inference, for each token, its logit ${\ell}_g$ is formed by ${\ell}_g = {\ell}_u + s({\ell}_c - {\ell}_u)$, where ${\ell}_c$ is conditional logit, ${\ell}_u$ is unconditional logit, and $s$ is scale of the classifier-free guidance.

It is worth noting that all design choices discussed so far are largely inspired by previous works, for example, image tokenizer is borrowed from~\citep{ldm,vit-vqgan}, image generation is from~\citep{dit,pixart,vqgan}. A large portion of these techniques are well studied in diffusion models but little in AR models. Our work adapts these advanced designs collectively to AR-based visual generation models.

\subsection{Scale Up}  
Our model architecture is almost the same as Llama, which allows us to seamlessly adopt optimization techniques~\citep{rmsnorm,swiglu,rope} and training recipes~\citep{flashattention,deepspeed,megatron} in LLM community. As shown in Table~\ref{tab:model_scaling}, we scale the model size up to 3.1B parameters in this work. All models are implemented with PyTorch 2~\citep{pytorch2} and trained on 80GB A100 GPUs. For training the models with parameters below 1.4B, we directly use DDP, otherwise, we adopt PyTorch FSDP~\citep{fsdp} to optimize GPU memory usage.

\begin{table*}
\centering
\begin{tabular}{@{}lcccc@{}}
\toprule
\bf Model & \bf Parameters & \bf Layers & \bf Hidden Size & \bf Heads \\
\midrule
\our-B & 111M & 12 & 768 &  12 \\
\our-L & 343M & 24 & 1024 &  16 \\
\our-XL & 775M & 36 & 1280 &  20 \\
\our-XXL & 1.4B & 48 & 1536 &  24 \\
\our-3B & 3.1B & 24 & 3200 &  32 \\
\bottomrule
\end{tabular}
\caption{\textbf{Model sizes and architecture configurations of \our.} The configurations are following previous works~\citep{gpt2,Llama1,open_Llama_3b}.
}
\label{tab:model_scaling}
\vspace{-3mm}
\end{table*}

\subsection{Serving}
Autoregressive models have always suffered from its low inference speed. With the rapid development of large language models, advanced inference techniques~\citep{vllm,speculative,bitsandbytes} are proposed in the LLM community to optimize the inference speed.  

Similar to training, inference techniques developed in the LLM community can also be adopted to optimize our models. We verify the effectiveness of vLLM~\citep{vllm}, one of the most popular LLM serving frameworks, on our image generation methods. As shown in Table~\ref{tab:vllm}, 326\% - 414\% speedup is achieved compared to the baseline setting.

\section{Experiments}
\subsection{Image Tokenizer}
\paragraph{Training setup.} The training is on ImageNet~\citep{imagenet} train set, using the resolution of 256$\times$256 and random crop data augmentation. The image tokenizer model size is 72M and 70M when the downsample ratio is 16 and 8, respectively. All models are trained with the same settings: constant learning rate of $10^{-4}$, AdamW optimizer with $\beta_1=0.9$, $\beta_2=0.95$, weight decay$=0.05$, batch size of 128 and training epochs of 40. For the training losses, commitment loss weight is $0.25$ and adversarial loss weight is $0.5$. The adversarial loss is enabled after 20k training iterations. 

\paragraph{Evaluation metrics.} We use the popular ImageNet benchmark under the image resolution of 256 $\times$ 256. The image reconstruction quality is measured by r-FID, reconstruction-FID on 256$\times$256 ImageNet 50k validation set. The codebook usage is calculated as the percentage of used codes in the queue of size 65536 over the whole codebook size. We also report PSNR and SSIM as the metrics of reconstruction quality, following SDXL~\citep{sdxl}.

\begin{table}[h]
\centering
\begin{subtable}{.48\linewidth}
\centering
\begin{tabular}{c|cccc}
\toprule
dim & rFID$\downarrow$ & PSNR$\uparrow$ & SSIM$\uparrow$ & usage$\uparrow$ \\
\midrule
256 & 9.21 & 18.32 & 0.575 & 0.29\% \\
32 &  3.22 & 19.98 & 0.646 & 20.9\% \\
8 & 2.19 & 20.79 & 0.675  & 97.0\% \\
4 & 9.88 & 19.39 & 0.593 & 82.0\% \\
\bottomrule
\end{tabular}
\caption{\textbf{Codebook vector dimension.} Lower vector dimension (from 256 to 8) improves both reconstruction quality and codebook usage significantly.}
\end{subtable}
\hfill
\begin{subtable}{.48\linewidth}
\centering
\begin{tabular}{c|cccc}
\toprule
size & rFID$\downarrow$ & PSNR$\uparrow$ & SSIM$\uparrow$ & usage$\uparrow$ \\
\midrule
4096 & 3.02 & 19.99 & 0.643 & 100.0\% \\
8192 & 2.91 & 20.41 & 0.654 & 75.0\% \\
16384 & 2.19 & 20.79 & 0.675 & 97.0\% \\
32768 & 2.26 & 20.59 & 0.663 & 85.0\%\\
\bottomrule
\end{tabular}
\caption{\textbf{Codebook size.} Larger codebook size (from 4096 to 16384) benefits to the overall performance of image tokenizers.}
\end{subtable}
\vspace{-2mm}
\caption{\textbf{Ablation studies on codebook designs in image tokenizers.}. The evaluations are on 256$\times$256 ImageNet 50k validation set. The default setting is codebook vector dimension is 8, codebook size is 16384, downsample ratio is 16. }
\label{tab:token_design}
\vspace{-5mm}
\end{table}

\begin{table}[h]
\centering
\begin{tabular}{ccc|cccc}
\toprule
ratio & img size & tokens size & rFID$\downarrow$ & PSNR$\uparrow$ & SSIM$\uparrow$ & usage$\uparrow$   \\
\midrule
 & 256 & 256 (16$\times$16) & 2.19 & 20.79 & 0.675 & 97.0\%\\
16 & 384 & 576 (24$\times$24) & 0.94 & 21.94 & 0.726 & 97.0\%\\
& 512 & 1024 (32$\times$32) & 0.70 & 23.03 & 0.772 & 97.0\%\\
\midrule
& 256 & 1024 (32$\times$32) & 0.59 & 24.45 & 0.813 & 97.6\%\\
8 & 384 & 2304 (48$\times$48) & 0.37 & 25.63 & 0.852 & 97.6\%\\
& 512 & 4096 (64$\times$64) & 0.39 & 26.98 & 0.888 & 97.6\%\\
\bottomrule
\end{tabular}
\vspace{2mm} 
\caption{\textbf{Number of tokens to represent the image.} The number of tokens depends on downsample ratio and input image size. The reconstructed image is always resized to 256$\times$256 when evaluating on ImageNet 50k validation set. The default setting is codebook vector dimension is 8, codebook size is 16384.}
\label{tab:num_tokens}
\vspace{-5mm}
\end{table}

\paragraph{Effect of image codebook designs.} As shown in Table~\ref{tab:token_design}, when the codebook vector dimension is reduced from 256 to 32 to 8, much better reconstruction quality and codebook usage are consistently achieved. For codebook size, a larger size from 4096 to 16384 benefits the overall performance. These observations are consistent with previous works~\citep{vit-vqgan, magvit2}.

\paragraph{Effect of number of tokens to represent the image.} Table~\ref{tab:num_tokens} studies the effect of image token number on image reconstruction quality. Using the same image tokenizer, for example, downsample ratio as 16, representing an image with only 256 tokens (16$\times$16) is not sufficient for good reconstruction quality, and increasing the number of tokens to 576 (24$\times$24) could largely improve the image quality from 2.43 to 0.99 rFID.

\begin{table}[t]
\centering
\begin{tabular}{m{4mm}m{24mm}cc|ccc|ccc}
\toprule
\multirow{2}{*}{ratio} & 
\multirow{2}{*}{method} & 
\multirow{2}{*}{dim} & 
\multirow{2}{*}{size} & 
\multicolumn{3}{c|}{ImageNet} & 
\multicolumn{3}{c}{COCO}\\
& & & & rFID$\downarrow$ & PSNR$\uparrow$ & SSIM$\uparrow$ & rFID$\downarrow$ & PSNR$\uparrow$ & SSIM$\uparrow$   \\
\midrule
 \multirow{4}{*}{16} & VQGAN & 256 & 1024 & 8.30 & 19.51 & 0.614 & 16.95 & 19.08 & 0.613\\
& VQGAN & 256 & 16384 & 4.99 & 20.00 & 0.629 & 12.29 & 19.57 & 0.630\\
& MaskGIT & 256 & 1024 & 2.28 & - & - & - & - & - \\
& \textbf{Ours} & 8 & 16384 & 2.19 & 20.79 & 0.675 & 8.11 & 20.42 & 0.678 \\
\midrule
\multirow{4}{*}{8} & VQGAN\textsuperscript{oim.} & 4 & 256 & 1.44 & 22.63 & 0.737 & 6.58 & 22.289 & 0.744\\
& VQGAN\textsuperscript{oim.} & 4 & 16384 & 1.19 & 23.38 & 0.762 & 5.89 & 23.08 & 0.771\\
& ViT-VQGAN & 32 & 8192 & 1.28 & - & - & - & - & -\\
& \textbf{Ours} & 8 & 16384 & 0.59 & 24.45 & 0.813 & 4.19 & 24.20 & 0.822\\
\midrule
\multirow{3}{*}{8} & SD-VAE\textsuperscript{ukn.} & 4 & - & 0.74 & 25.68 & 0.820 & 4.45 & 25.41 & 0.831\\
& SDXL-VAE\textsuperscript{ukn.} & 4 & - & 0.68 & 26.04 & 0.834 & 4.07 & 25.76 & 0.845\\
& OAI-Decoder\textsuperscript{ukn.} & 4 & - & 0.81 & 24.43 & 0.786 & 4.59 & 24.19 & 0.800\\
\bottomrule
\end{tabular}
\vspace{2mm} 
\caption{\textbf{Comparisons with other image tokenizers.} The evaluations are on 256$\times$256 ImageNet 50k validation set and COCO 5k \texttt{val2017} set. All models are trained on ImageNet except ``oim.'' is on OpenImage, ``ukn.'' is unknown training data. }
\label{tab:tokenizer}
\vspace{-5mm}
\end{table}

\paragraph{Comparisons with other image tokenizers.}
We compare with other image tokenizers, including VQGAN~\citep{vqgan}, MaskGIT~\citep{maskgit}, ViT-VQGAN~\citep{vit-vqgan}.  As shown in Table~\ref{tab:tokenizer}, our tokenizer outperforms previous image tokenizers. We also evaluate our tokenizer on COCO val2017~\citep{coco} of 256 $\times$ 256 image resolution to verify the image reconstruction quality, since COCO images contain more complex scenes. The comparison results are consistent with those in ImageNet validation set. This shows our tokenizer is a generalizable image tokenizer for both object-centric and scene-centric images.

Importantly, our tokenizer is competitive to \textit{continuous} latent space representation, such as SD VAE~\citep{ldm}, SDXL VAE~\citep{sdxl}, and Consistency Decoder from OpenAI~\citep{consistencydecoder}, which are widely used in diffusion models. This shows that \textit{discrete} representation in the image tokenizer is no longer the bottleneck of the image reconstruction.

\subsection{Class-conditional Image Generation}

\paragraph{Training setup.} Our benchmark is the popular 256 $\times$ 256 ImageNet. All models are trained with the similar settings: base learning rate of $10^{-4}$ per 256 batch size, AdamW optimizer with $\beta_1=0.9$, $\beta_2=0.95$, $\text{weight decay}=0.05$, gradient clipping of 1.0. The dropout is always 0.1 for input token embedding, attention module and FFN module.
The class condition embedding dropout for classifier-free guidance is 0.1.

\textbf{Precomputing image codes.} To accelerate the model training, we use the image tokenizer to precompute image codes before training. To achieve the similar effect of random crop data augmentation, we extract image codes of ten crops of the original image. During training, we randomly select one copy code from the ten augmentations.

\paragraph{Evaluation metrics.} We use Fréchet inception distance (FID)~\citep{fid} as the main metric. We also report Inception Score (IS)~\citep{inception_score}, sFID~\citep{sfid} and Precision/Recall~\citep{precision_recall} as secondary metrics. All evaluations are implemented using ADM’s TensorFlow scripts~\citep{adm} for fair comparisons.

\begin{table}[!t]
\centering
\begin{tabular}{ll|cccc}
\toprule
image token & model & FID$\downarrow$ & IS$\uparrow$ & Precision$\uparrow$ & Recall$\uparrow$ \\
\midrule
\multirow{5}{*}{\shortstack[l]{image size: 256$\times$256\\ tokens: 256 (16$\times$16)\\ rFID: 2.19}}  & B & 8.69 & 124.43 & 0.78 & 0.46 \\
 & L & 4.21 & 200.00 & 0.82 & 0.50\\
 & XL & 3.39 & 227.08 & 0.81 & 0.54\\
 & XXL & 3.09 & 253.60 & 0.82 & 0.52 \\
 & 3B & 3.06 & 279.71 & 0.84 & 0.53\\
\midrule
\multirow{5}{*}{\shortstack[l]{image size: 384$\times$384 \\ tokens: 576 (24$\times$24) \\ rFID: 0.94}}  & B & 12.89 & 92.44 & 0.73 &  0.48\\
 & L & 5.01 & 167.31 & 0.78 & 0.52\\
 & XL & 3.42 & 202.93 & 0.79 & 0.56\\
 & XXL & 2.89 & 236.21 & 0.80 & 0.56 \\
 & 3B & 2.61 & 251.90 & 0.80 & 0.56\\
\bottomrule
\end{tabular}
\vspace{2mm}
\caption{\textbf{The effect of image tokens on image generation.} The generated image is always resized to 256$\times$256 when evaluating on ImageNet benchmark. We compare all models after training 50 epochs. The inference setting is cfg = 1.75, top-k = 0 (all), top-p = 1.0, temperature = 1.0 for all experiments.}
\label{tab:img_tokens}
\vspace{-5mm}
\end{table}

\begin{figure}[!t]
    \centering
    \begin{subfigure}[b]{0.48\textwidth}
        \centering
        \includegraphics[width=\textwidth]{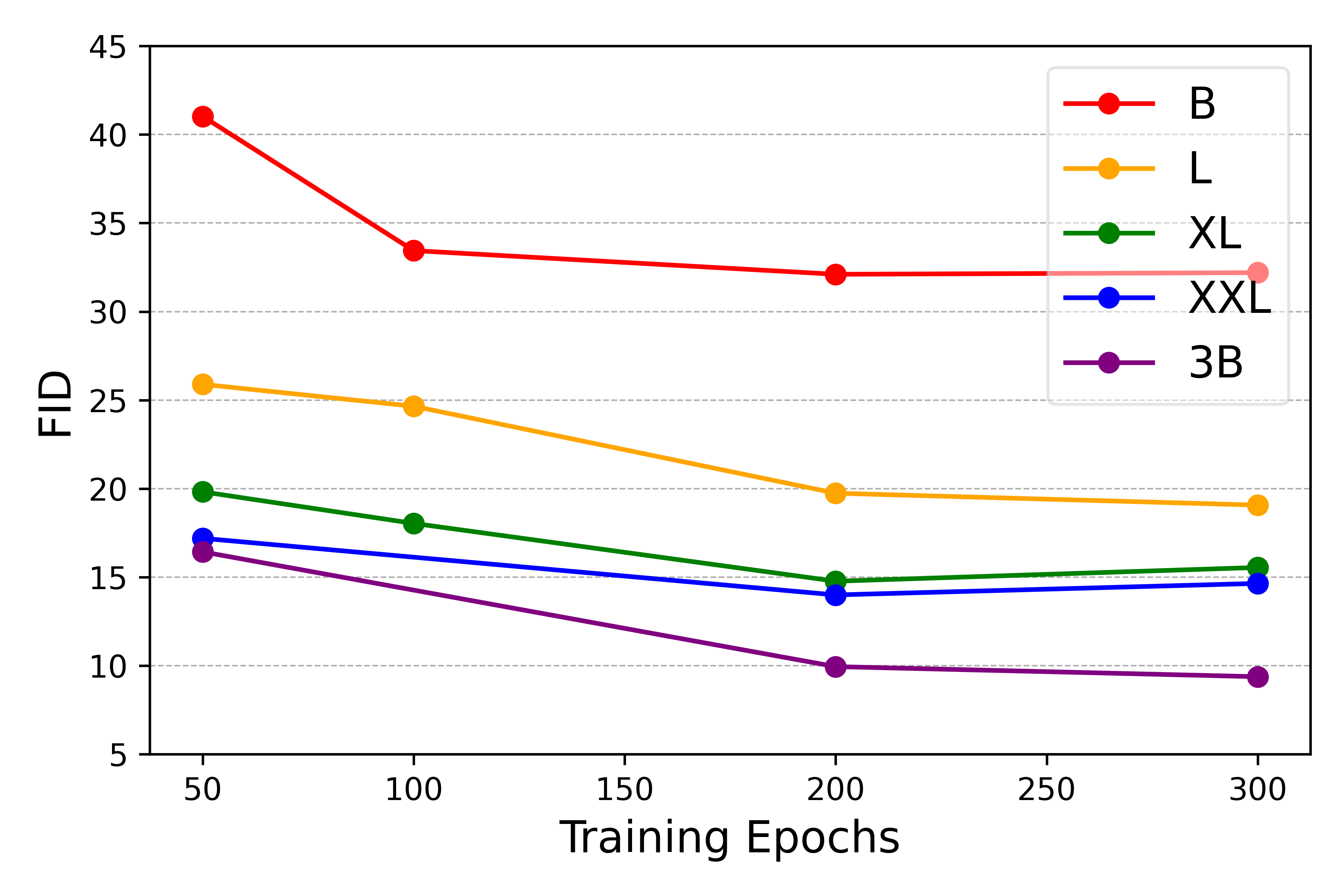}
        \vspace{-6mm}
        \caption{without classifier-free guidance}
        \label{fig:scaling_fig1}
    \end{subfigure}
    \begin{subfigure}[b]{0.48\textwidth}
        \centering
        \includegraphics[width=\textwidth]{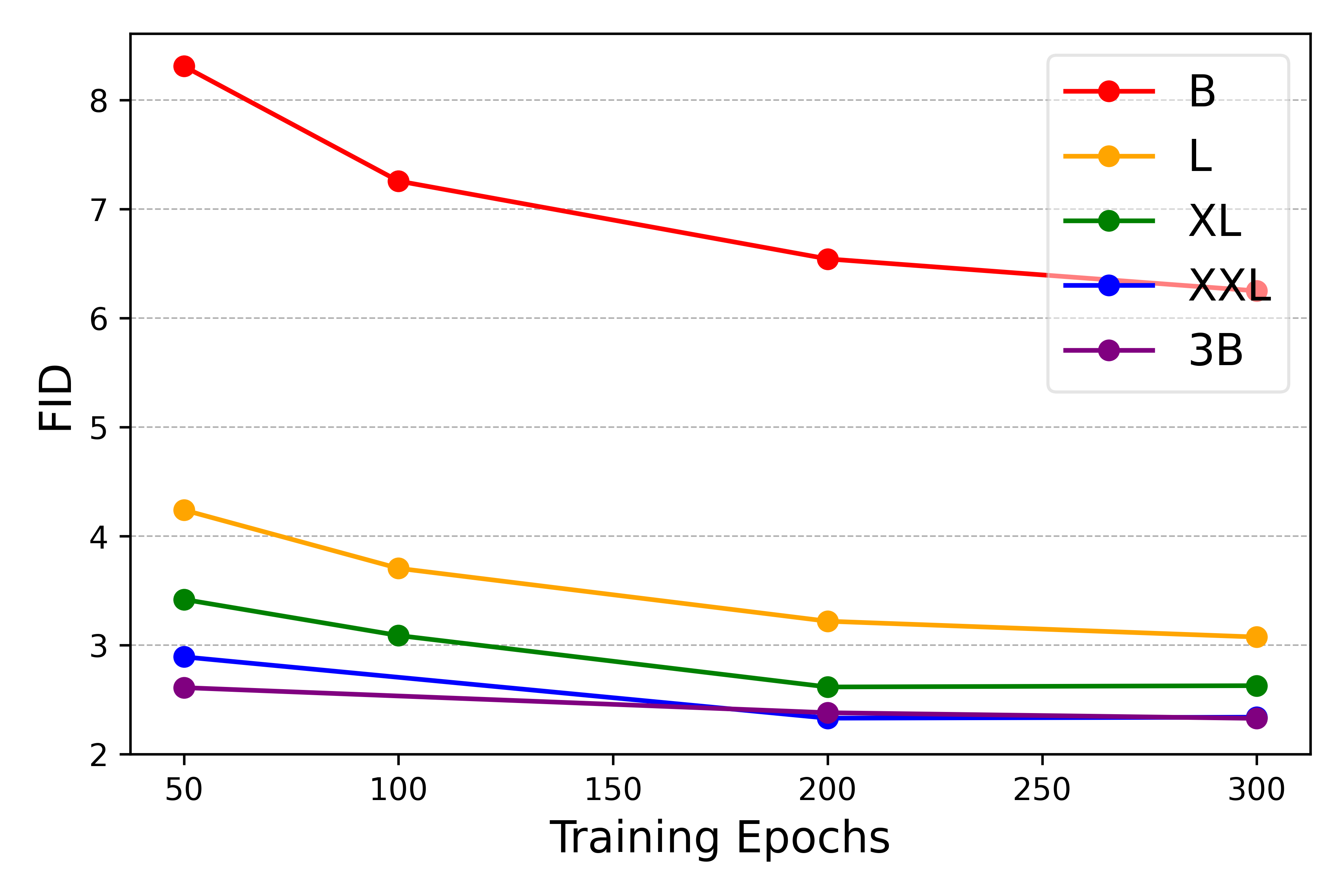}
         \vspace{-6mm}
        \caption{with classifier-free guidance}
        \label{fig:scaling_fig2}
    \end{subfigure}
\caption{\textbf{Scaling model size.} We show FID of 256$\times$256 ImageNet benchmark over training epochs. Scaling model size brings consistent improvement on FID during the whole training process. More detailed evaluation metrics are in Appendix. 
}
\label{fig:scaling}
\vspace{-5mm}
\end{figure}

\begin{figure}[!t]
    \centering
    \begin{subfigure}[b]{0.45\textwidth}
        \centering
        \includegraphics[width=\textwidth]{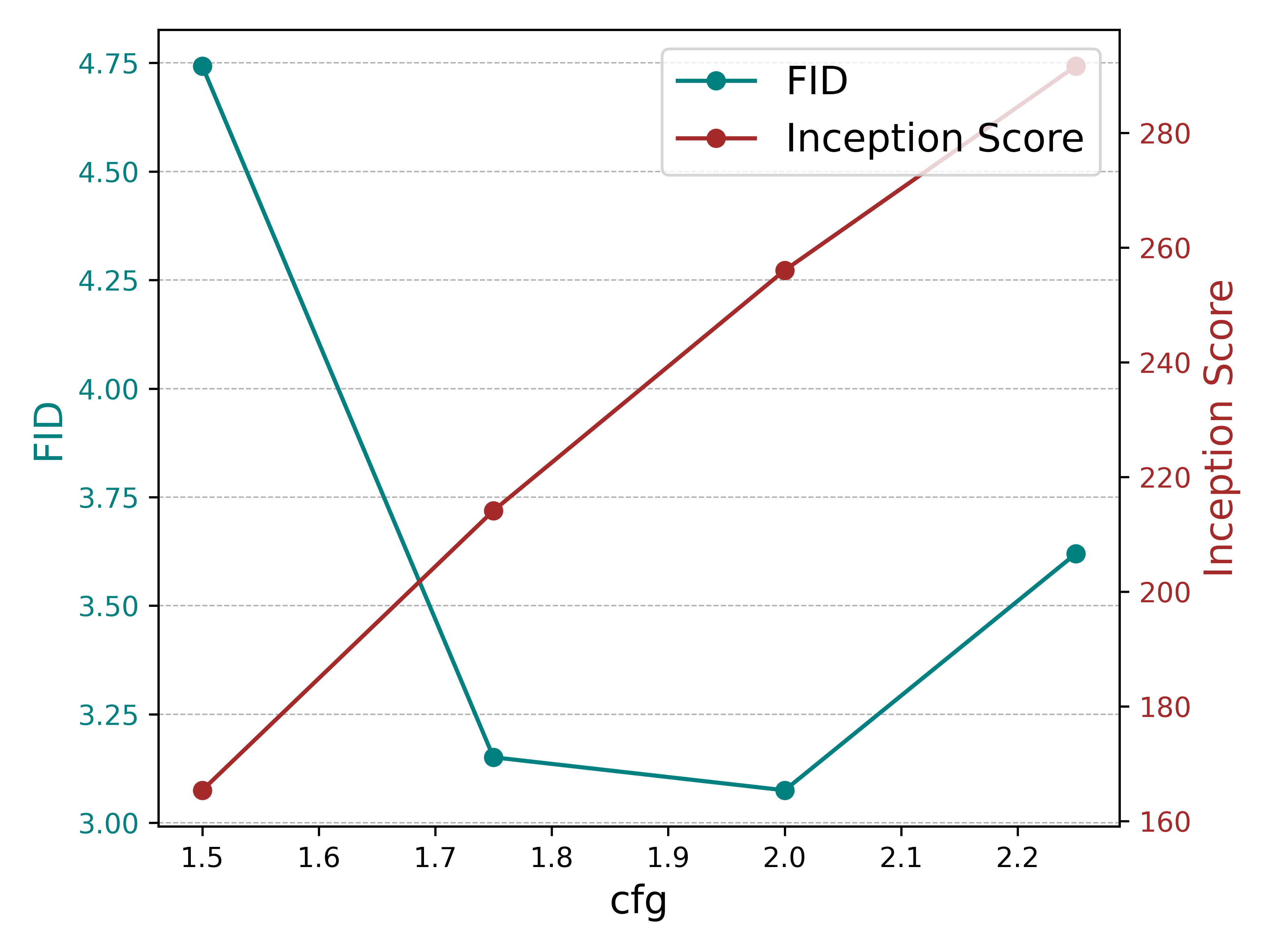}
         \vspace{-6mm}
        \caption{classifier-free guidance}
        \label{fig:sample_cfg}
    \end{subfigure}
    \hspace{2mm}
    \begin{subfigure}[b]{0.45\textwidth}
        \centering
        \includegraphics[width=\textwidth]{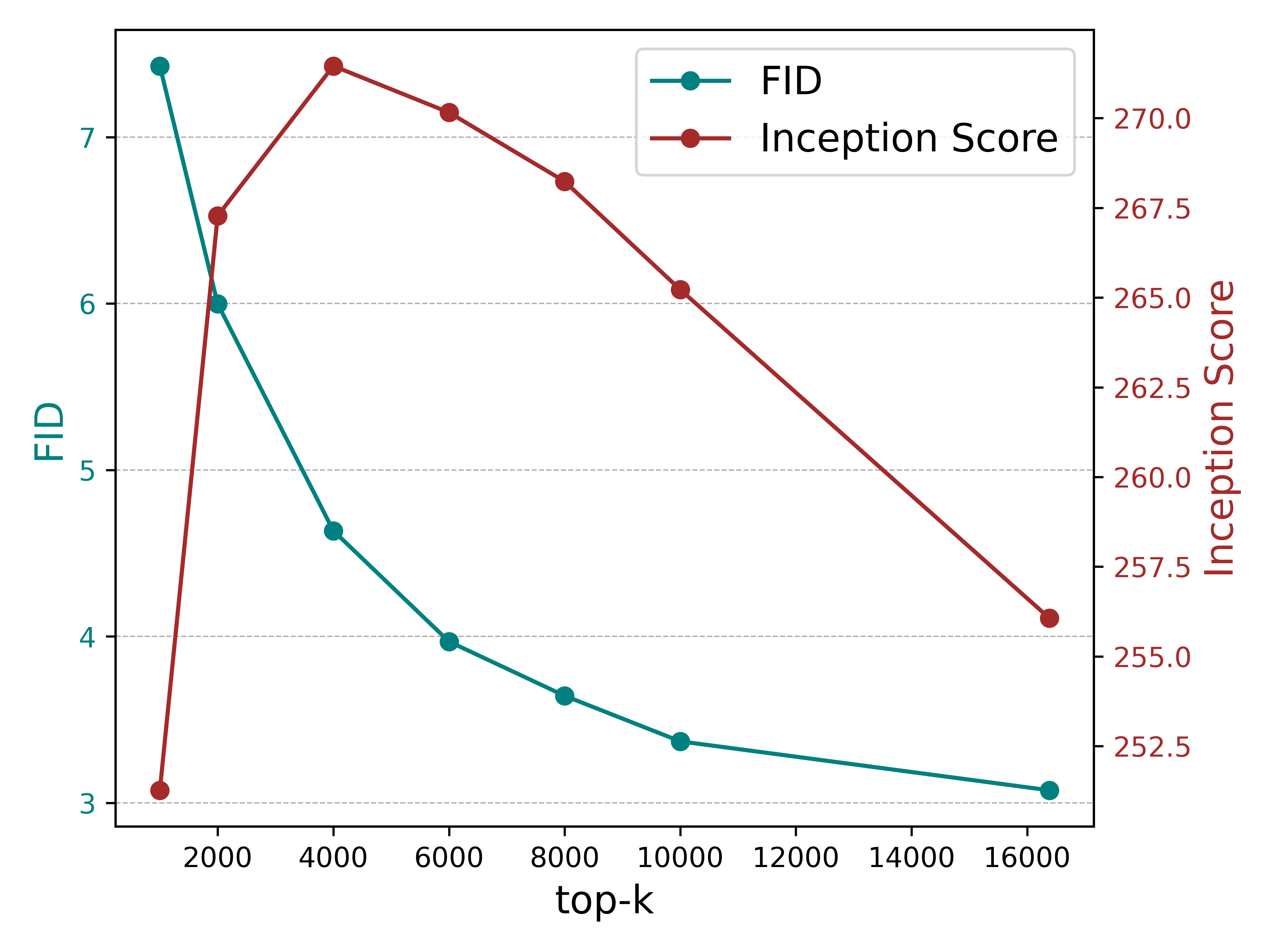}
         \vspace{-6mm}
        \caption{top-k sampling}
        \label{fig:sample_topk}
    \end{subfigure}
\caption{\textbf{The effect of sampling configuration.} We show FID and Inception Score of 256×256 ImageNet benchmark over different sampling configurations. The model is \our-L, and the default setting is cfg = 2.0, top-k = 0 (all), top-p = 1.0, temperature = 1.0.
}
\label{fig:sampling}
\vspace{-5mm}
\end{figure}

\paragraph{Effect of image tokens.}  Although increasing the image tokens brings better image reconstruction quality, it is not strongly correlated to image generation quality. As shown in Table~\ref{tab:img_tokens}, when the model parameter is smaller than 1B, 256 (16$\times$16) tokens bring better image generation performance than 576 (24$\times$24). This shows the synergistic effect of scaling up model parameters and token numbers. Nevertheless, fewer image tokens would limit the image generation performance, for example, 256 (16$\times$16) tokens limit the FID at 3.06 FID, while 576 (24$\times$24) could further improve the FID to a lower value.

\begin{table}[t]
\centering
\begin{tabular}{c|lc|cccc}
\toprule
Type & Model & \#Para. & FID$\downarrow$ & IS$\uparrow$ & Precision$\uparrow$ & Recall$\uparrow$  \\
\midrule
\multirow{3}{*}{GAN}   & BigGAN~\citep{biggan}  & 112M   & 6.95  & 224.5       & 0.89 & 0.38 \\
 & GigaGAN~\citep{gigagan}  & 569M    & 3.45  & 225.5       & 0.84 & 0.61  \\
 & StyleGan-XL~\citep{stylegan-xl} & 166M    & 2.30  & 265.1       & 0.78 & 0.53   \\
\midrule
\multirow{4}{*}{Diffusion} & ADM~\citep{adm}  & 554M       & 10.94 & 101.0        & 0.69 & 0.63    \\
 & CDM~\citep{cdm}   & $-$       & 4.88  & 158.7       & $-$  & $-$   \\
 & LDM-4~\citep{ldm} & 400M     & 3.60  & 247.7       & $-$  & $-$  \\
 & DiT-XL/2~\citep{dit}  & 675M  & 2.27  & 278.2       & 0.83 & 0.57   \\
\midrule
\multirow{2}{*}{Mask.} & MaskGIT~\citep{maskgit}  & 227M   & 6.18  & 182.1        & 0.80 & 0.51  \\
 & MaskGIT-re~\citep{maskgit} & 227M\    & 4.02  & 355.6        & $-$ & $-$ \\
\midrule
\multirow{7}{*}{AR} & VQGAN~\citep{vqgan} & 227M & 18.65 & 80.4         & 0.78 & 0.26    \\
 & VQGAN~\citep{vqgan}    & 1.4B   & 15.78 & 74.3   & $-$  & $-$     \\
 & VQGAN-re~\citep{vqgan}  & 1.4B  & 5.20  & 280.3  & $-$  & $-$     \\
 & ViT-VQGAN~\citep{vit-vqgan} & 1.7B & 4.17  & 175.1  & $-$  & $-$        \\
 & ViT-VQGAN-re~\citep{vit-vqgan}& 1.7B  & 3.04  & 227.4  & $-$  & $-$     \\
 & RQTran.~\citep{rq}       & 3.8B  & 7.55  & 134.0  & $-$  & $-$     \\
 & RQTran.-re~\citep{rq}    & 3.8B & 3.80  & 323.7  & $-$  & $-$    \\
\midrule
\multirow{7}{*}{AR} & \our-B (cfg=2.00) & 111M & 5.46 & 193.61 & 0.83 & 0.45\\
 & \our-L  (cfg=2.00) & 343M & 3.07 & 256.06 & 0.83 & 0.52 \\
 & \our-XL  (cfg=1.75) & 775M & 2.62 & 244.08 & 0.80 & 0.57 \\
 & \our-XXL (cfg=1.75) & 1.4B & 2.34 & 253.90 & 0.80 & 0.59 \\
 & \our-3B (cfg=1.65) & 3.1B & 2.18 & 263.33 & 0.81 & 0.58 \\
 & \our-3B (cfg=1.75) & 3.1B & 2.32 & 280.10 & 0.82 & 0.56 \\
 & \our-3B (cfg=2.00) & 3.1B & 2.81 & 311.59 & 0.84 & 0.54 \\

\bottomrule
\end{tabular}
\vspace{3mm}
\caption{\textbf{Model comparisons on class-conditional ImageNet 256$\times$256 benchmark}. Metrics include Fréchet inception distance (FID), inception score (IS), precision and recall.
``$\downarrow$'' or ``$\uparrow$'' indicate lower or higher values are better.
``-re'' means using rejection sampling. ``cfg'' means using classifier-free guidance. More detailed results are in Appendix.
}
\label{tab:main}
\vspace{-5mm}
\end{table}

\paragraph{Effect of model size.} 
We train our models across five model sizes (B, L, XL, XXL, 3B) and evaluate their performance with and without classifier-free guidance. Figure~\ref{fig:scaling} illustrates how FID changes as both the model sizes and the training epochs increase. Notable improvements in FID are observed when scaling the model from \our-B to \our-XXL. Further scaling to 3B yields only marginal improvements. A plausible explanation for this phenomenon could be the limitation in dataset size: ImageNet~\citep{imagenet} comprises approximately only 1 million images, expanding the dataset or using stronger data augmentation could potentially lead to further improvements.

\paragraph{Effect of classifier-free guidance (CFG).} First, as shown in Figure~\ref{fig:scaling}, using classifier-free guidance can significantly enhance the visual quality across all model sizes. Moreover, Figure~\ref{fig:sample_cfg} illustrates that the model achieves optimal FID at CFG = 2.0 and further increasing CFG would deteriorate FID, which is consistent with previous findings~\citep{adm}. Additionally, the increment in CFG results in a trade-off between diversity and fidelity, as evidenced by increased precision and decreased recall, demonstrated in Table~\ref{tab:app_3}.

\paragraph{Effect of top-k sampling.} As shown in Figure~\ref{fig:sample_topk}, a small top-k value is not beneficial for FID and IS. Increasing top-k continuously improves FID but decreases IS, which trades off fidelity for diversity. We observe a similar trend when changing the parameter of top-p and temperature in sampling. Since FID is our main metric, we use maximum value as the default top-k value, which is the whole codebook size.

\paragraph{Comparisons with other image generation methods.} 
In Table~\ref{tab:main}, we compare with popular image generation models, including GAN~\citep{biggan,gigagan,stylegan-xl}, Diffusion models~\citep{adm,cdm,ldm,dit}, and masked-prediction models~\citep{maskgit}. Our models exhibit competitive performance in all metrics of FID, IS, Precision and Recall. Notably, our 3B model outperforms the popular diffusion models LDM~\citep{ldm}, DiT~\citep{dit}. This shows that vanilla autoregressive models can serve as the basis of advanced image generation systems.

When comparing with autoregressive models~\citep{vqgan,vit-vqgan,rq}, our model outperforms all previous models at different levels of model parameters. This benefits from better designs of image tokenizers and better scalability of image generation models. We hope our simple and effective implementation will serve as a solid baseline and help facilitate future research in autoregressive models for image generations.

\subsection{Text-conditional Image Generation}

\paragraph{Training setup.} We adopt a two-stage training strategy. In stage I, the model is trained on a 50M subset of LAION-COCO~\citep{laion_coco} with the image resolution 256$\times$256. In Stage II, the model is fine-tuned on 10M internal high aesthetic quality images with the image resolution 512$\times$512. Examples of training data are shown in the Appendix. The maximum length of text token embedding is set to 120, and left padding is used to enable batch processing. The text condition embedding dropout for classifier-free guidance is 0.1. All models are trained with similar settings: model parameters of 775M, base learning rate of $10^{-4}$ per 256 batch size, AdamW optimizer with $\beta_1=0.9$, $\beta_2=0.95$, $\text{decay}=0.05$, gradient clipping of 1.0.

\textbf{Precomputing image codes and text embeddings.}
We use pre-trained FLAN-T5 XL~\citep{flan_t5} to precompute text embedding of the image captions. For image code, we only extract image codes of the original image center crop in text-conditional models training.

\paragraph{Fine-tune image tokenizer.} Before two-stage training for text-conditional image generation models, we first fine-tune the image tokenizer on the joint of 50M LAION-COCO and 10M internal high aesthetic quality data.

\begin{figure}[!t]
\begin{center}
\begin{tabular}{c}
\includegraphics[width=0.99\textwidth]{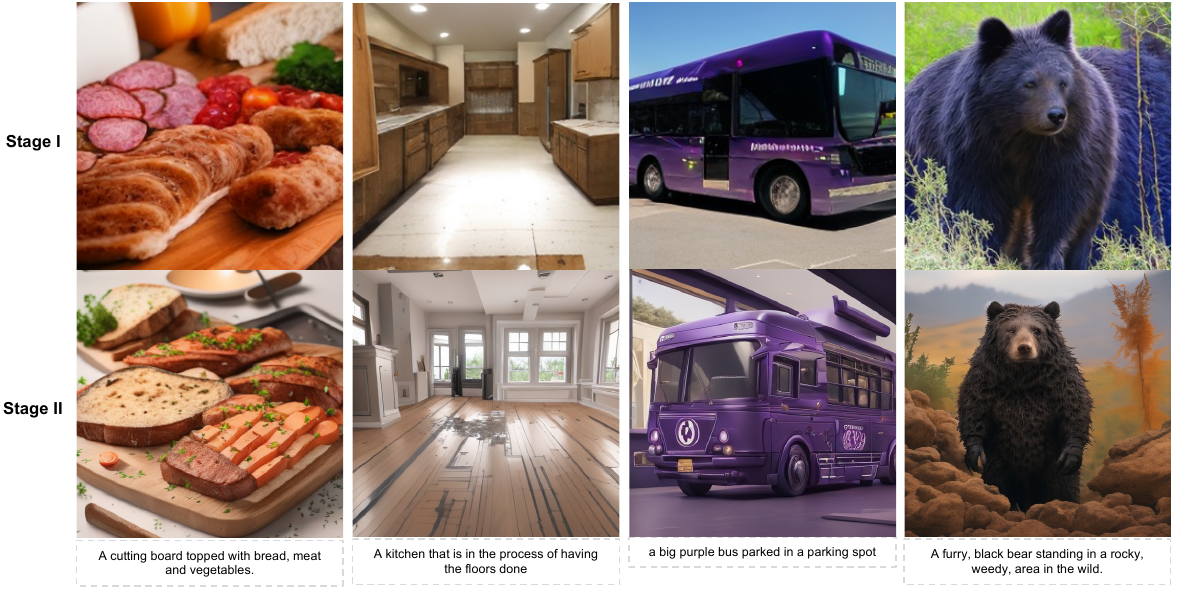}
\end{tabular}
\end{center}
\vspace{-3mm}
\caption{\textbf{Visualization of two-stage training of text-conditional image generation models.} Comparisons of generated images by models after stage I training and stage II training. The text prompts are from COCOPrompts. }
\label{fig:stage1_and_2}
\end{figure}

\paragraph{Visualizations.} In Figure~\ref{fig:stage1_and_2}, we select text prompts from COCOPrompts~\citep{coco} to generate images using models after stage I training and stage II training.
After stage I training, the model captures the text-image alignment, while its ability to represent image details is not clear.  Stage II training improves the visual aesthetic quality by a significant margin. We explain this improvement comes from two aspects: high aesthetic quality images shift the domain, and high image resolution brings better visual details. We notice that further increasing the image resolution to 1024$\times$1024 could bring better visual quality, and we leave it for future research.

More visualizations on PartiPrompts~\citep{parti} are in Appendix. PartiPrompts have more longer captions than COCOPrompts, and our model demonstrates competitive performance in text-image alignment for long caption image generation tasks.

\paragraph{Limitation.} Due to the training data and model parameters, our text-conditional models have several limitations, such as text rendering errors, counting errors, and common misconceptions. These problems are promising to be mitigated when more training data and computation resources are available in the future.

\subsection{Inference Speed} 
We verify the effectiveness of vLLM~\citep{vllm} serving framework on our methods. Since our models use the same architecture as Llama, which is already supported by vLLM, we can seamlessly adopt its implementation. As shown in Table~\ref{tab:vllm}, we achieve 326\% - 414\% speedup compared to the baseline setting in the models from 111M to 1.4B parameters. Please note that the baseline setting has already integrated KV-Cache technique. In the 3B model, its head size 100 is not supported by PagedAttention in vLLM.

\begin{table*}
\centering
\begin{tabular}{@{}ccccc@{}}
\toprule
model &  parameters & baseline (sec) & vllm (sec) & speed-up ratio \\
\midrule
B & 111M & 7.80 & 2.39 &  326\% \\
L & 343M & 13.72 & 3.48 &  380\% \\
XL & 775M & 19.76 & 4.84 &  408\% \\
XXL & 1.4B & 26.38 & 6.36 &  414\% \\
\bottomrule
\end{tabular}
\caption{\textbf{Optimized inference speed by vLLM serving framework.} The inference time is for a batch 16 images (generating 8 images with classifier-free guidance). The image resolution is 384$\times$384 for all models.}
\label{tab:vllm}
\end{table*}

\section{Related Work}

\paragraph{Visual generation.}
Generative adversarial network (GAN)~\citep{gan,biggan,stylegan,gigagan} is the first representative visual generation method in deep learning era. To improve the distribution coverage, several likelihood-based methods are proposed. Diffusion models~\citep{ddpm,scorebased,ddim,adm} view image generation as the reverse diffusion process from noises to images. Masked-prediction models~\citep{maskgit,muse,magvit,magvit2} apply language model BERT-style~\citep{bert} by learning to predict masked tokens. Instead, autoregressive models~\citep{vqgan,dalle1,parti} leverage GPT-style~\citep{gpt1} to predict the next token in a sequence. To ease the modeling and improve the generation quality, these methods always introduce the image tokenization process~\citep{vae, vqvae} to convert pixel space to semantic space.

\paragraph{Multimodal foundation models.}
Recently, vision-and-language models~\citep{llava, minigpt4, instructblip, kosmos-2, gpt4roi, groma} have achieved versatile visual understanding through visual instruction tuning~\citep{llava, minigpt4}. However, unifying the understanding and generation in multimodal models is still in its early stages. Most existing methods~\citep{emu_baai, emu2_baai, dreamllm, seed} try to collaborate a pre-trained diffusion model with existing models, rather than utilizing a unified next-token prediction paradigm. These methods need sophisticated designs to connect the two parts with distinct training paradigms, which makes scaling up challenging. Pioneering methods~\citep{unified-io, unified-io2,lvm,gemini,chameleon} attempt to incorporate image generation into LLM using an autoregressive approach and achieve promising results. They do not specifically focus on demonstrating that a plain autoregressive approach can serve as a scalable image generator, which is our main argument in this work.

\section{Conclusion}
In this work, we delve into vanilla autoregressive models for scalable image generation. By reexamining their image tokenizers, image generation models and training data, our class-conditional models outperform the popular diffusion models, and our text-conditional models demonstrate competitive performance of visual quality and text alignment. 
\newpage

\bibliographystyle{llamagen}
\bibliography{reference}

\newpage
\appendix

\section{Examples of Image-Text Pair Data} 

\begin{figure}[!t]
\begin{center}
\begin{tabular}{c}
\includegraphics[width=0.99\textwidth]{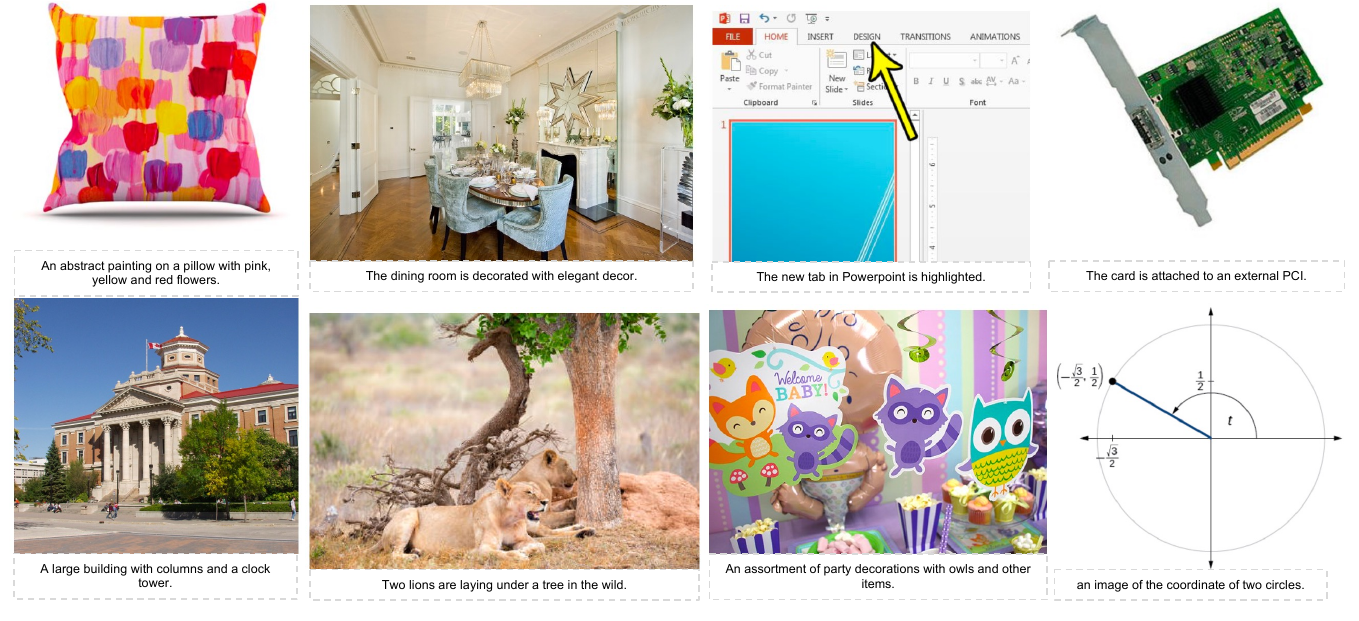}
\end{tabular}
\end{center}
\vspace{-3mm}
\caption{\textbf{Examples of stage I training data: 50M subset of LAION-COCO.} 
The short caption is its original caption (generated from BLIP~\citep{blip}).}
\label{fig:stage1}
\end{figure}

\paragraph{Training stage I: 50M subset of LAION-COCO~\citep{laion_coco}.} The original dataset has 600M image-text pair.  We filter these images by valid image URL, aesthetic score, watermark score, CLIP image-text similarity score and image size. The remaining images are about 50M. Some examples are shown in Figure~\ref{fig:stage1}.

\begin{figure}[!t]
\begin{center}
\begin{tabular}{c}
\includegraphics[width=0.99\textwidth]{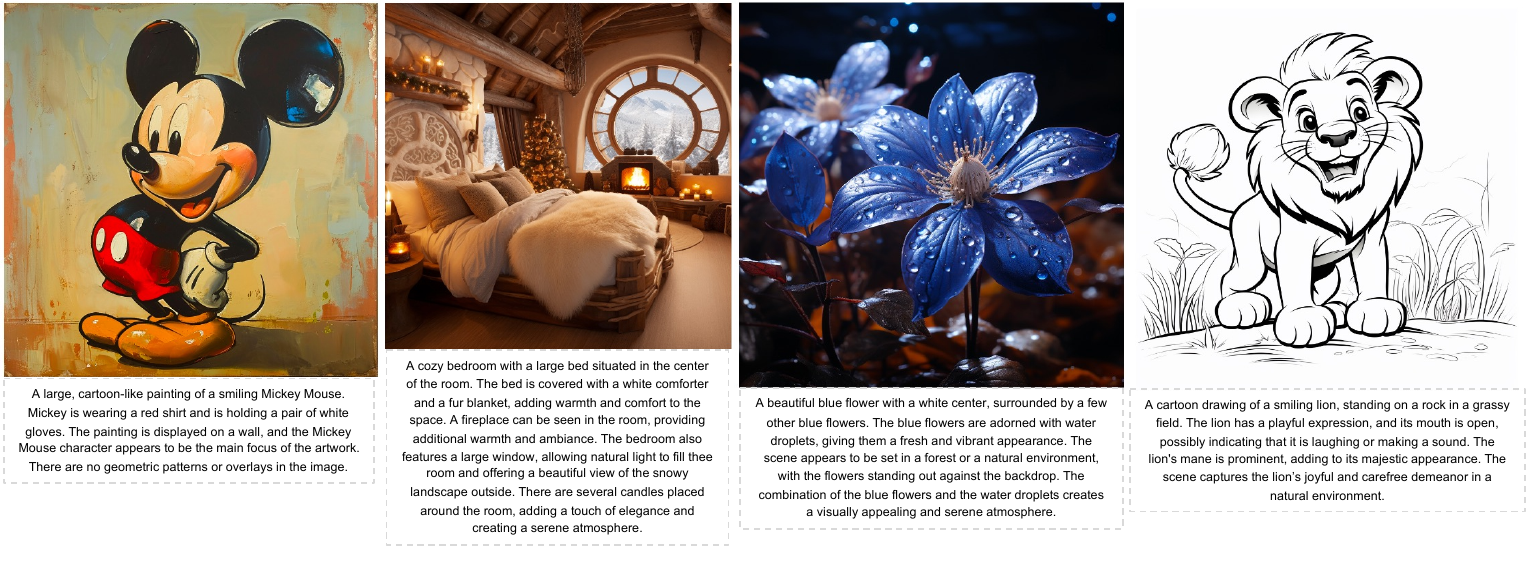}
\end{tabular}
\end{center}
\vspace{-3mm}
\caption{\textbf{Examples of stage II training data: 10M internal high aesthetic quality images.} The long caption is generated from LLaVA.
}
\label{fig:stage2}
\end{figure}

\paragraph{Training stage II: 10M internal high aesthetic quality images.} 
Each image is provided a long caption by LLaVA~\citep{llava} using the prompt of ``Describe this image in as much detail as possible''. Some examples are shown in Figure~\ref{fig:stage2}. We notice that the first sentence of the long caption is always a summary description of its image, so we use it as the short caption to augment the training of text-conditional image generation models. 

\section{More Results on ImageNet Benchmark}

We provide more detailed performance on ImageNet 256$\times$256 benchmark in Table~\ref{tab:app_1}~\ref{tab:app_2}~\ref{tab:app_3}. The generated image is always resized to 256$\times$256 when evaluating.

\newpage

\textcolor{white}{...}

\textcolor{white}{...}

\textcolor{white}{...}

\begin{table}[!t]
\centering
\begin{tabular}{cccr|ccccc}
\toprule
Model & \#Para. & epochs & cfg  & FID$\downarrow$  &  IS$\uparrow$ & sFID$\downarrow$ & Pre.$\uparrow$ & Rec.$\uparrow$   \\
\midrule
B & 111M & 50 & no & 31.352 & 39.576 & 8.749 & 0.568 & 0.614\\
B & 111M & 50 & 1.50 & 11.984 & 95.400 & 7.335 & 0.738 & 0.517\\
B & 111M & 50 & 1.75 & 8.690 & 124.435 & 7.165 & 0.789 & 0.469\\
B & 111M & 50 & 2.00 & 7.390 & 153.974 & 7.250 & 0.832 & 0.417\\
B & 111M & 50 & 2.25 & 7.220 & 178.281 & 7.489 & 0.861 & 0.384\\ 
B & 111M & 50 & 2.50 & 7.824 & 197.511 & 7.857 & 0.882 & 0.349\\
\midrule
B & 111M & 300 & no & 26.262 & 48.072 & 9.216 & 0.593 & 0.616\\
B & 111M & 300 & 1.50 & 8.738 & 120.602 & 7.668 & 0.751 & 0.535\\
B & 111M & 300 & 1.75 & 6.116 & 159.123 & 7.364 & 0.799 & 0.492\\
B & 111M & 300 & 2.00 & 5.464 & 193.613 & 7.503 & 0.839 & 0.457\\
B & 111M & 300 & 2.25 & 5.641 & 220.720 & 7.668 & 0.863 & 0.411\\ 
B & 111M & 300 & 2.50 & 6.390 & 246.565 & 8.041 & 0.883 & 0.382\\
\midrule
L & 343M & 50 & no & 21.812 & 59.179 & 8.772 & 0.616 & 0.640\\
L & 343M & 50 & 1.50 & 5.781 & 153.792 & 7.096 & 0.774 & 0.555\\
L & 343M & 50 & 1.75 & 4.218 & 200.001 & 7.015 & 0.824 & 0.509\\
L & 343M & 50 & 2.00 & 4.317 & 242.112 & 7.077 & 0.859 & 0.468\\
\midrule
L & 343M & 300 & no & 13.452 & 82.289 & 8.324 & 0.656 & 0.638\\
L & 343M & 300 & 1.50 & 4.079 & 198.504 & 8.157 & 0.800 & 0.552\\
L & 343M & 300 & 1.75 & 3.805 & 248.280 & 8.487 & 0.833 & 0.515\\
L & 343M & 300 & 2.00 & 4.407 & 288.170 & 8.871 & 0.858 & 0.481\\
\midrule
XL & 775M & 50 & no & 19.417 & 66.196 & 8.911 & 0.610 & 0.665\\
XL & 775M & 50 & 1.50 & 4.808 & 172.170 & 7.298 & 0.767 & 0.585\\
XL & 775M & 50 & 1.75 & 3.391 & 227.081 & 7.022 & 0.812 & 0.542\\
XL & 775M & 50 & 2.00 & 3.642 & 268.779 & 7.244 & 0.846 & 0.502\\
\midrule
XXL & 1.4B & 50 & no & 16.822 & 74.888 & 9.285 & 0.628 & 0.660\\
XXL & 1.4B & 50 & 1.50 & 3.844 & 195.527 & 7.496 & 0.781 & 0.577\\
XXL & 1.4B & 50 & 1.75 & 3.094 & 253.609 & 7.305 & 0.825 & 0.529\\
XXL & 1.4B & 50 & 2.00 & 3.644 & 296.521 & 7.410 & 0.857 & 0.511\\
\midrule
3B & 3.1B & 50 & no & 13.581 & 87.902 & 7.781 & 0.648 & 0.666\\
3B & 3.1B & 50 & 1.50 & 3.050 & 222.330 & 6.489 & 0.801 & 0.575\\
3B & 3.1B & 50 & 1.75 & 3.063 & 279.716 & 6.686 & 0.843 & 0.538\\
3B & 3.1B & 50 & 2.00 & 4.212 & 325.150 & 7.027 & 0.869 & 0.492\\
\midrule
validation data &  & & & 1.684 & 231.811 & 3.692 & 0.752 & 0.671\\
\bottomrule
\end{tabular}
\vspace{3mm}
\caption{\textbf{Detailed performance on class-conditional ImageNet 256$\times$256 benchmark}. The generated image is 256$\times$256. All experiments  use the sampling configuration of top-k = 0 (all), top-p = 1.0, temperature = 1.0. 
}
\label{tab:app_1}
\end{table}

\begin{table}[!t]
\centering
\begin{tabular}{cccr|ccccc}
\toprule
Model & \#Para. & epochs & cfg  & FID$\downarrow$  & sFID$\downarrow$ & IS$\uparrow$ & Pre.$\uparrow$ & Rec.$\uparrow$   \\
\midrule
B & 111M & 50 & no & 41.025 & 30.788 & 9.825 & 0.523 & 0.605\\
B & 111M & 50 & 1.50 & 18.276 & 69.337 & 7.557 & 0.677 & 0.534 \\
B & 111M & 50 & 1.75 & 12.899 & 92.447 & 6.900 & 0.738 & 0.487\\
B & 111M & 50 & 2.00 & 10.029 & 116.372 & 6.562 & 0.787 & 0.443\\
B & 111M & 50 & 2.25 & 8.674 & 136.621 & 6.428 & 0.818 & 0.413\\
B & 111M & 50 & 2.50 & 8.309 & 154.719 & 6.599 & 0.843 & 0.376\\
B & 111M & 50 & 2.75 & 8.391 & 168.629 & 6.708 & 0.860 & 0.345\\
\midrule
B & 111M & 100 & no & 33.442 & 37.528 & 9.872 & 0.536 & 0.609\\
B & 111M & 100 & 1.50 & 15.629 & 77.247 & 7.632 & 0.698 & 0.529\\
B & 111M & 100 & 1.75 & 10.676 & 104.581 & 6.960 & 0.754 & 0.490\\
B & 111M & 100 & 2.00 & 8.298 & 128.941 & 6.671 & 0.795 & 0.452\\
B & 111M & 100 & 2.25 & 7.256 & 152.502 & 6.510 & 0.827 & 0.416\\
B & 111M & 100 & 2.50 & 7.151 & 172.677 & 6.517 & 0.850 & 0.390\\
\midrule
B & 111M & 200 & no & 32.105 & 37.993 & 10.144 & 0.559 & 0.618\\
B & 111M & 200 & 1.50 & 12.206 & 90.783 & 7.531 & 0.716 & 0.534\\
B & 111M & 200 & 1.75 & 8.535 & 118.399 & 7.024 & 0.766 & 0.503\\
B & 111M & 200 & 2.00 & 6.951 & 146.077 & 6.784 & 0.808 & 0.459\\
B & 111M & 200 & 2.25 & 6.542 & 167.825 & 6.695 & 0.833 & 0.428\\
B & 111M & 200 & 2.50 & 6.632 & 188.157 & 6.811 & 0.853 & 0.393\\
\midrule
B & 111M & 300 & no & 32.196 & 39.877 & 11.838 & 0.570 & 0.611\\
B & 111M & 300 & 1.50 & 12.012 & 95.553 & 8.897 & 0.725 & 0.528\\
B & 111M & 300 & 1.75 & 8.012 & 127.957 & 8.088 & 0.778 & 0.498\\
B & 111M & 300 & 2.00 & 6.437 & 157.173 & 7.487 & 0.814 & 0.456\\
B & 111M & 300 & 2.25 & 6.092 & 182.538 & 7.244 & 0.845 & 0.416\\
B & 111M & 300 & 2.50 &  6.249 & 203.886 & 6.981 & 0.861 & 0.389\\
B & 111M & 300 & 2.75 & 6.803 & 220.708 & 6.928 & 0.876 & 0.357\\
\midrule
L & 343M & 50 & no & 25.889 & 48.053 & 9.612 & 0.570 & 0.655\\
L & 343M & 50 & 1.50 & 7.905 & 123.830 & 7.381 & 0.732 & 0.569\\
L & 343M & 50 & 1.75 & 5.018 & 167.310 & 6.786 & 0.784 & 0.524\\
L & 343M & 50 & 2.00 & 4.240 & 206.739 & 6.483 & 0.825 & 0.491\\
L & 343M & 50 & 2.25 & 4.589 & 238.890 & 6.325 & 0.850 & 0.451\\
\midrule
L & 343M & 100 & no & 24.654 & 53.166 & 10.497 & 0.594 & 0.645\\
L & 343M & 100 & 1.50 & 6.934 & 138.852 & 7.910 & 0.748 & 0.569\\
L & 343M & 100 & 1.75 & 4.321 & 188.536 & 7.068 & 0.802 & 0.528\\
L & 343M & 100 & 2.00 & 3.705 & 228.305 & 6.701 & 0.839 & 0.490\\
L & 343M & 100 & 2.25 & 4.054 & 263.864 & 6.407 & 0.858 & 0.460\\
\midrule
L & 343M & 200 & no & 19.742 & 61.715 & 7.286 & 0.601 & 0.667\\
L & 343M & 200 & 1.50 & 4.929 & 158.546 & 6.066 & 0.759 & 0.588\\
L & 343M & 200 & 1.75 & 3.249 & 209.372 & 5.927 & 0.805 & 0.544\\
L & 343M & 200 & 2.00 & 3.220 & 250.697 & 5.879 & 0.841 & 0.512\\
L & 343M & 200 & 2.25 & 3.939 & 288.217 & 6.076 & 0.865 & 0.479\\
\midrule
L & 343M & 300 & no & 19.070 & 64.349 & 8.668 & 0.607 & 0.670\\
L & 343M & 300 & 1.50 & 4.743 & 165.381 & 6.740 & 0.758 & 0.596\\
L & 343M & 300 & 1.75 & 3.151 & 214.152 & 6.310 & 0.803 & 0.552\\
L & 343M & 300 & 2.00 & 3.075 & 256.067 & 6.088 & 0.832 & 0.522\\
L & 343M & 300 & 2.25 & 3.620 & 291.695 & 6.122 & 0.854 & 0.493\\
\midrule
validation data &  & & & 1.684 & 231.811 & 3.692 & 0.752 & 0.671\\
\bottomrule
\end{tabular}
\vspace{3mm}
\caption{\textbf{Detailed performance on class-conditional ImageNet 256$\times$256 benchmark}. The generated image is 384$\times$384 and is resized to 256$\times$256 when evaluating on ImageNet. All experiments  use the sampling configuration of top-k = 0 (all), top-p = 1.0, temperature = 1.0. 
}
\label{tab:app_2}
\end{table}

\begin{table}[!t]
\centering
\begin{tabular}{cccr|ccccc}
\toprule
Model & \#Para. & epochs & cfg  & FID$\downarrow$  & sFID$\downarrow$ & IS$\uparrow$ & Pre.$\uparrow$ & Rec.$\uparrow$   \\
\midrule
XL & 775M & 50 & no & 19.820 & 61.363 & 8.067 & 0.601 & 0.669\\
XL & 775M & 50 & 1.50 & 5.231 & 154.249 & 6.284 & 0.746 & 0.592\\
XL & 775M & 50 & 1.75 & 3.420 & 202.939 & 6.090 & 0.796 & 0.560\\
XL & 775M & 50 & 2.00 & 3.238 & 245.680 & 6.023 & 0.826 & 0.529\\
\midrule
XL & 775M & 100 & no & 18.037 & 69.879 & 8.388 & 0.616 & 0.665\\
XL & 775M & 100 & 1.50 & 4.563 & 173.749 & 6.591 & 0.759 & 0.588\\
XL & 775M & 100 & 1.75 & 3.089 & 225.856 & 6.157 & 0.804 & 0.551\\
XL & 775M & 100 & 2.00 & 3.105 & 267.608 & 6.001 & 0.833 & 0.531\\
\midrule
XL & 775M & 200 & no & 14.772 & 80.826 & 6.840 & 0.620 & 0.681\\
XL & 775M & 200 & 1.50 & 3.388 & 193.477 & 5.753 & 0.771 & 0.603\\
XL & 775M & 200 & 1.75 & 2.617 & 245.465 & 5.652 & 0.811 & 0.566\\
XL & 775M & 200 & 2.00 & 2.859 & 285.900 & 5.758 & 0.840 & 0.527\\
\midrule
XL & 775M & 300 & no & 15.549 & 79.157 & 7.049 & 0.616 & 0.689\\
XL & 775M & 300 & 1.50 & 3.479 & 194.448 & 5.816 & 0.763 & 0.606\\
XL & 775M & 300 & 1.75 & 2.629 & 244.085 & 5.594 & 0.807 & 0.579\\
XL & 775M & 300 & 2.00 & 2.785 & 286.875 & 5.567 & 0.836 & 0.542\\
\midrule
XXL & 1.4B & 50 & no & 17.195 & 74.123 & 8.689 & 0.605 & 0.681\\
XXL & 1.4B & 50 & 1.50 & 4.363 & 178.228 & 6.818 & 0.758 & 0.600\\
XXL & 1.4B & 50 & 1.75 & 2.893 & 236.210 & 6.263 & 0.805 & 0.564\\
XXL & 1.4B & 50 & 2.00 & 3.049 & 285.390 & 6.053 & 0.842 & 0.522\\
\midrule
XXL & 1.4B & 200 & no & 13.997 & 86.776 & 8.178 & 0.637 & 0.684\\
XXL & 1.4B & 200 & 1.50 & 3.137 & 207.870 & 6.060 & 0.774 & 0.605\\
XXL & 1.4B & 200 & 1.75 & 2.331 & 262.995 & 5.714 & 0.816 & 0.579\\
XXL & 1.4B & 200 & 2.00 & 2.678 & 304.631 & 5.587 & 0.840 & 0.545\\
\midrule
XXL & 1.4B & 300 & no & 14.648 & 86.328 & 8.687 & 0.628 & 0.681\\
XXL & 1.4B & 300 & 1.50 & 3.295 & 202.586 & 6.476 & 0.770 & 0.626\\
XXL & 1.4B & 300 & 1.75 & 2.340 & 253.906 & 5.977 & 0.809 & 0.596\\
XXL & 1.4B & 300 & 2.00 & 2.523 & 295.374 & 5.736 & 0.836 & 0.559\\
\midrule
3B & 3.1B & 50 & no & 16.431 & 72.622 & 7.217 & 0.611 & 0.677\\
3B & 3.1B & 50 & 1.50 & 3.472 & 191.979 & 5.955 & 0.768 & 0.600\\
3B & 3.1B & 50 & 1.75 & 2.611 & 251.903 & 6.167 & 0.807 & 0.568\\
3B & 3.1B & 50 & 2.00 & 3.222 & 300.887 & 5.764 & 0.847 & 0.523\\
\midrule
3B & 3.1B & 200 & no & 9.949 & 108.083 & 7.088 & 0.667 & 0.672\\
3B & 3.1B & 200 & 1.50 & 2.400 & 237.683 & 5.548 & 0.794 & 0.600\\
3B & 3.1B & 200 & 1.65 & 2.264 & 268.180 & 5.426 & 0.817 & 0.581\\
3B & 3.1B & 200 & 1.75 & 2.381 & 286.091 & 5.390 & 0.828 & 0.569\\
3B & 3.1B & 200 & 2.00 & 3.011 & 321.563 & 5.514 & 0.851 & 0.538\\
\midrule
3B & 3.1B & 300 & no & 9.380 & 112.877 & 8.242 & 0.685 & 0.668\\
3B & 3.1B & 300 & 1.50 & 2.388 & 233.246 & 6.145 & 0.798 & 0.601\\
3B & 3.1B & 300 & 1.60 & 2.216 & 251.338 & 6.002 & 0.811 & 0.584\\
3B & 3.1B & 300 & 1.65 & 2.189 & 263.334 & 5.965 & 0.819 & 0.581\\
3B & 3.1B & 300 & 1.75 & 2.329 & 280.104 & 5.818 & 0.828 & 0.566\\
3B & 3.1B & 300 & 1.80 & 2.370 & 287.452 & 5.825 & 0.834 & 0.570\\
3B & 3.1B & 300 & 2.00 & 2.816 & 311.597 & 5.845 & 0.848 & 0.544\\
\midrule
validation data &  & & & 1.684 & 231.811 & 3.692 & 0.752 & 0.671\\
\bottomrule
\end{tabular}
\vspace{3mm}
\caption{\textbf{Detailed performance on class-conditional ImageNet 256$\times$256 benchmark}. The generated image is 384$\times$384 and is resized to 256$\times$256 when evaluating on ImageNet. All experiments  use the sampling configuration of top-k = 0 (all), top-p = 1.0, temperature = 1.0. 
}
\label{tab:app_3}
\end{table}

\begin{minipage}[t]{0.49\textwidth}
    \begin{figure}[H]
    \centering
    \includegraphics[width=\textwidth]{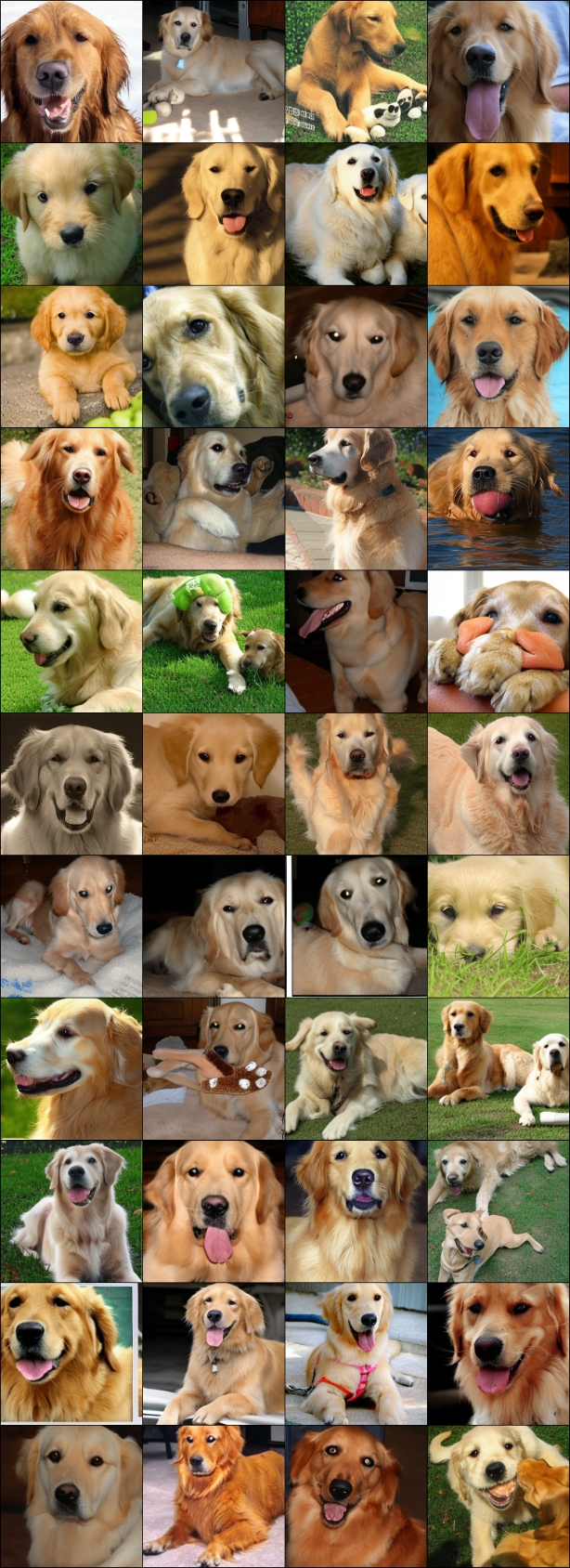}
    \caption{384$\times$384 \our-3B samples. \\ Classifier-free guidance scale = 4.0 \\
    Class label = "golden retriever" (207)}
    \label{fig:c2i_207}
    \end{figure}
\end{minipage}
\hfill
\begin{minipage}[t]{0.49\textwidth}
    \begin{figure}[H]
    \centering
    \includegraphics[width=\textwidth]{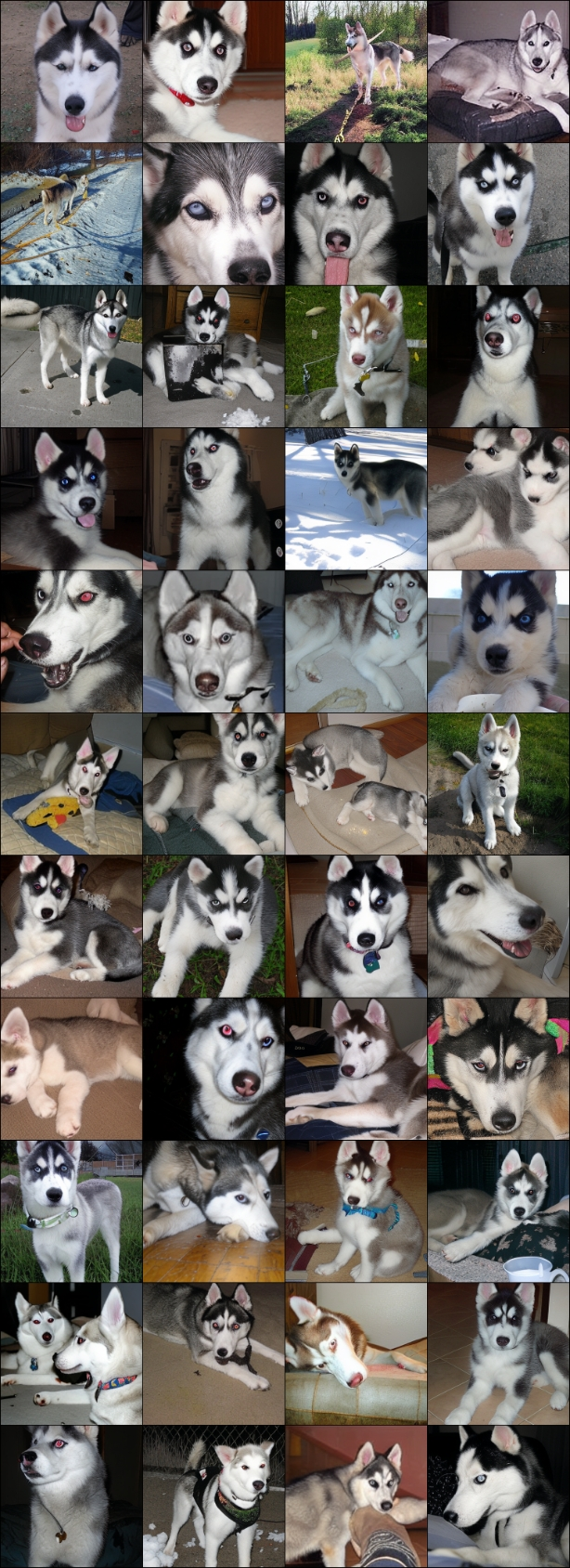}
    \caption{384$\times$384 \our-3B samples. \\
    Classifier-free guidance scale = 4.0 \\
    Class label = "husky " (250)}
    \label{fig:c2i_250}
    \end{figure}
\end{minipage}

\begin{minipage}[t]{0.49\textwidth}
    \begin{figure}[H]
    \centering
    \includegraphics[width=\textwidth]{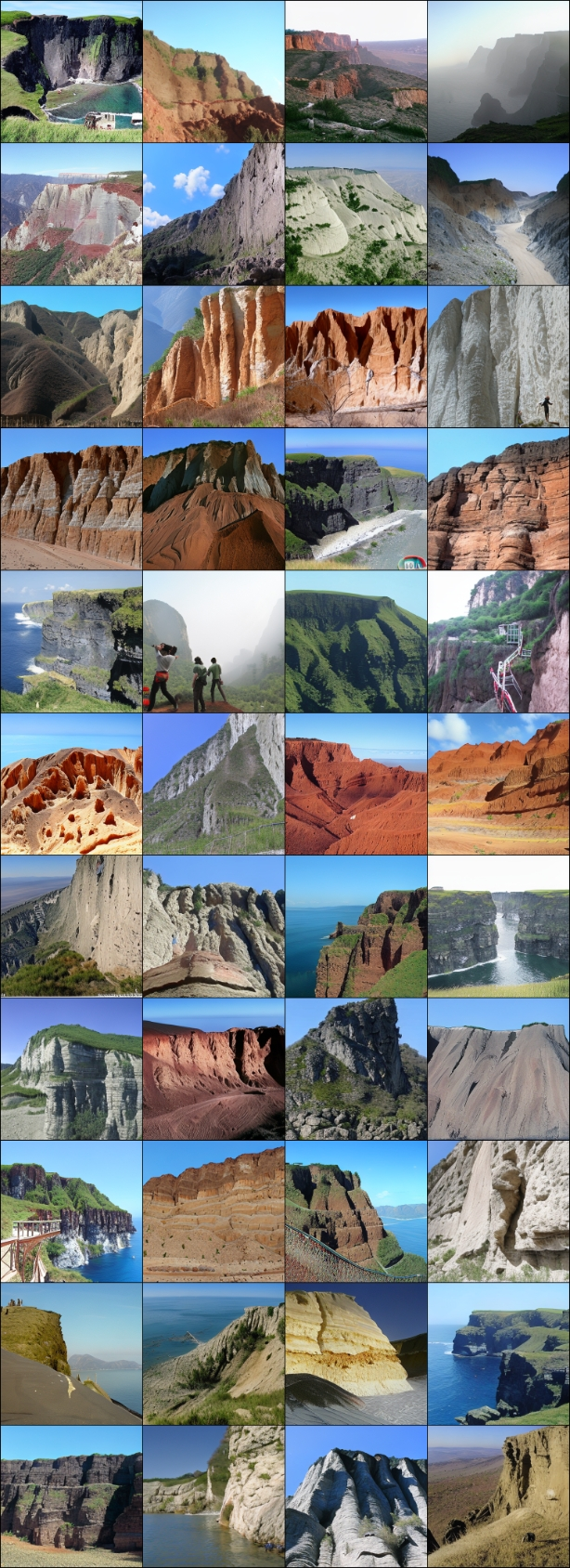}
    \caption{384$\times$384 \our-3B samples. \\ Classifier-free guidance scale = 4.0 \\
    Class label = "cliff drop-off" (972)}
    \label{fig:c2i_972}
    \end{figure}
\end{minipage}
\hfill
\begin{minipage}[t]{0.49\textwidth}
    \begin{figure}[H]
    \centering
    \includegraphics[width=\textwidth]{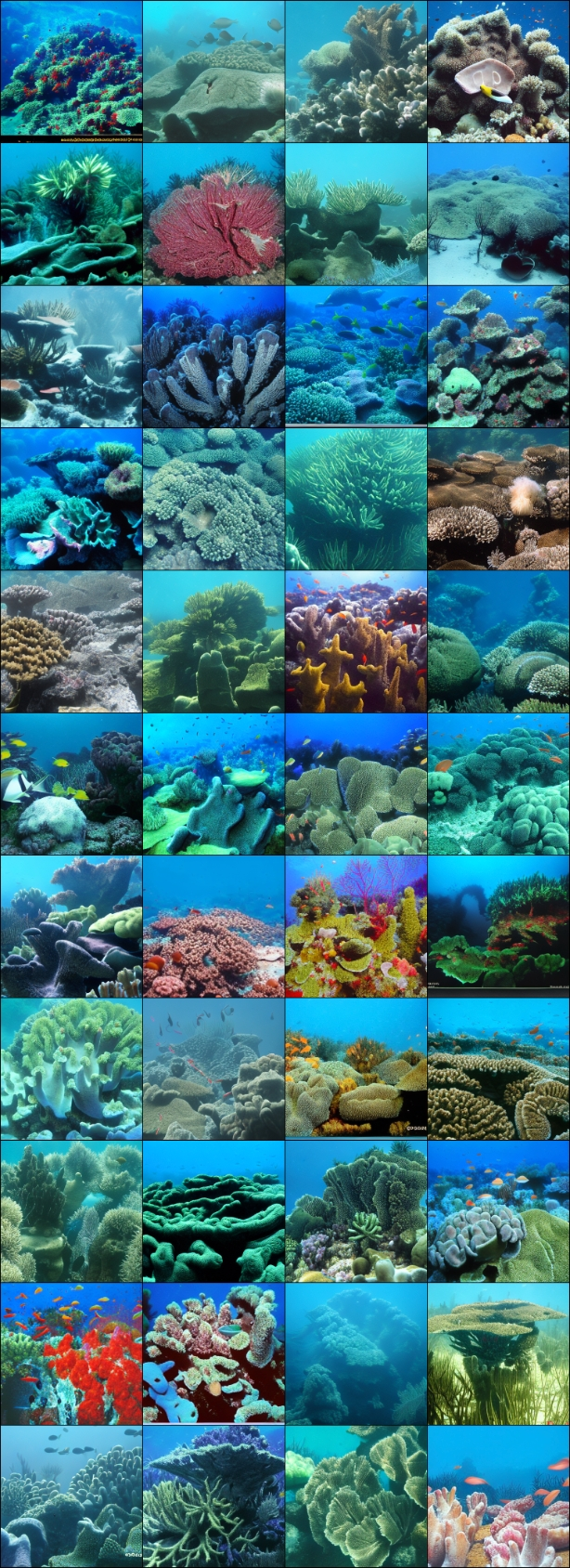}
    \caption{384$\times$384 \our-3B samples. \\
    Classifier-free guidance scale = 4.0 \\
    Class label = "coral reef" (973)}
    \label{fig:c2i_973}
    \end{figure}
\end{minipage}

\begin{minipage}[t]{0.49\textwidth}
    \begin{figure}[H]
    \centering
    \includegraphics[width=\textwidth]{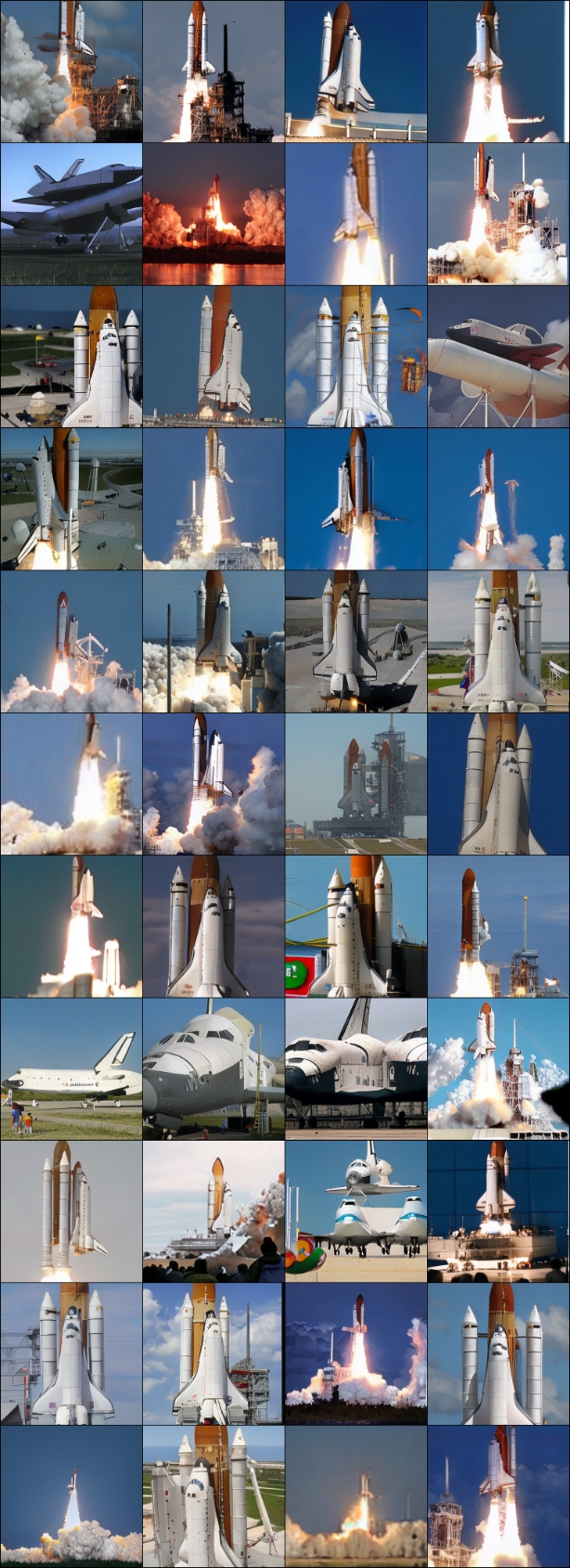}
    \caption{384$\times$384 \our-3B samples. \\ Classifier-free guidance scale = 4.0 \\
    Class label = "space shuttle" (812)}
    \label{fig:c2i_812}
    \end{figure}
\end{minipage}
\hfill
\begin{minipage}[t]{0.49\textwidth}
    \begin{figure}[H]
    \centering
    \includegraphics[width=\textwidth]{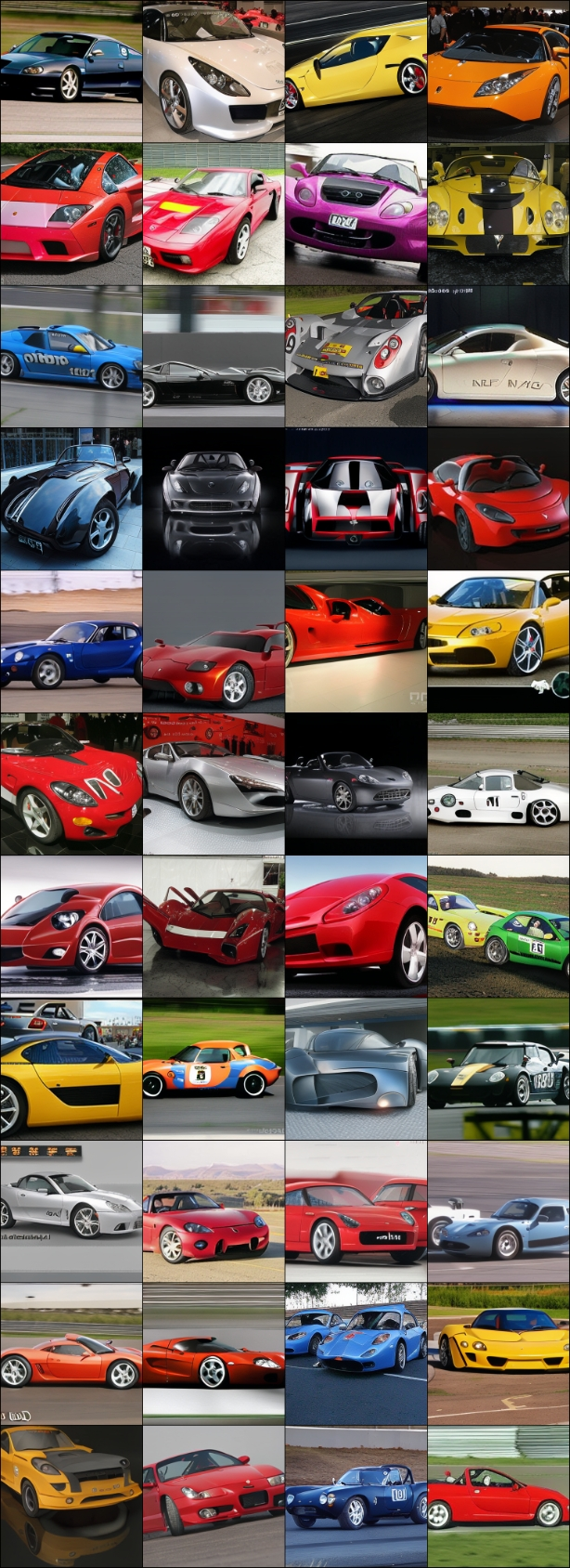}
    \caption{384$\times$384 \our-3B samples. \\
    Classifier-free guidance scale = 4.0 \\
    Class label = "sport car " (817)}
    \label{fig:c2i_817}
    \end{figure}
\end{minipage}

\begin{minipage}[t]{0.49\textwidth}
    \begin{figure}[H]
    \centering
    \includegraphics[width=\textwidth]{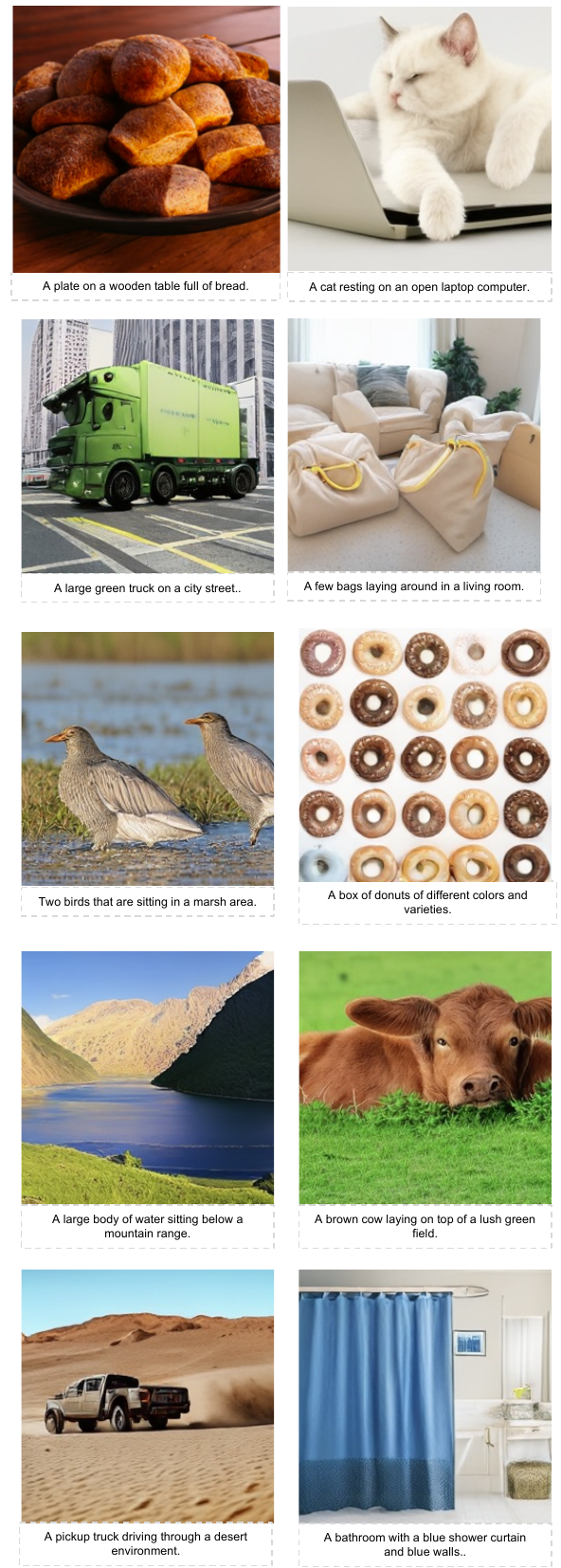}
    \caption{(Stage I) Text-conditional 256$\times$256 image generation on COCOPrompts.}
    \label{fig:stage1_coco}
    \end{figure}
\end{minipage}
\hfill
\begin{minipage}[t]{0.49\textwidth}
    \begin{figure}[H]
    \centering
    \includegraphics[width=\textwidth]{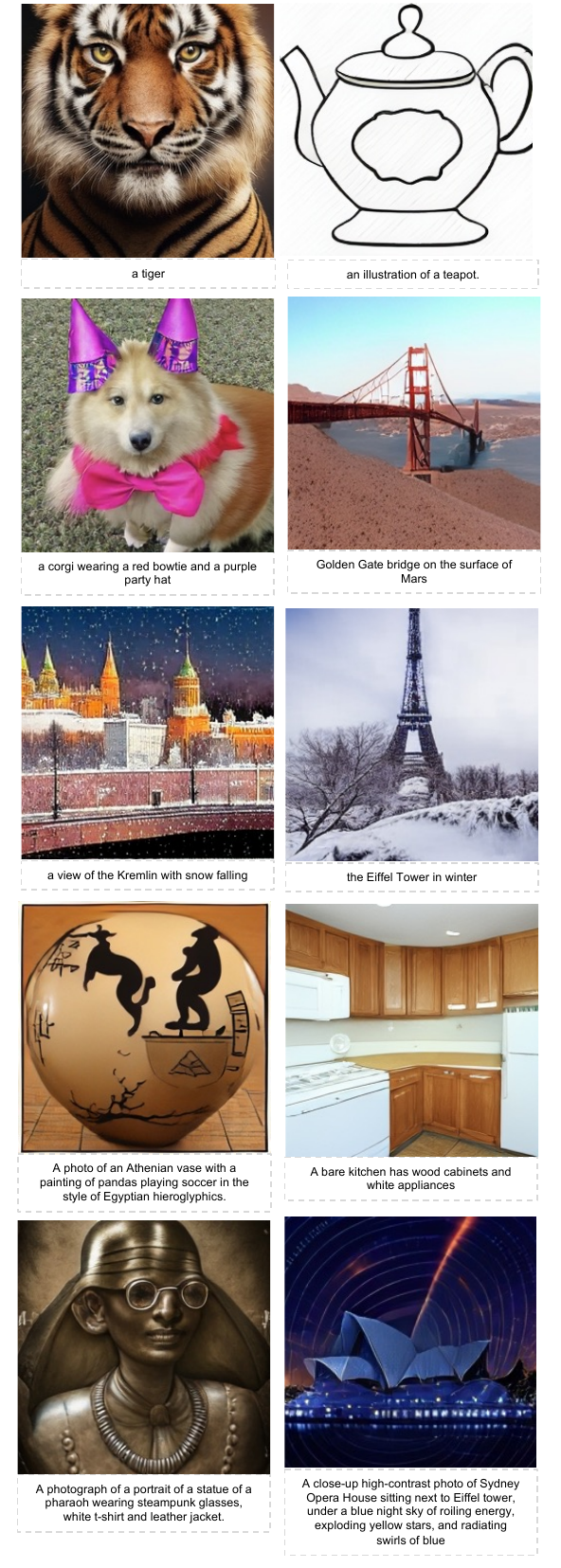}
    \caption{(Stage I) Text-conditional 256$\times$256 image generation on PartiPrompts.}
    \label{fig:stage1_parti}
    \end{figure}
\end{minipage}

\begin{figure}[!t]
\centering
\includegraphics[width=\textwidth]{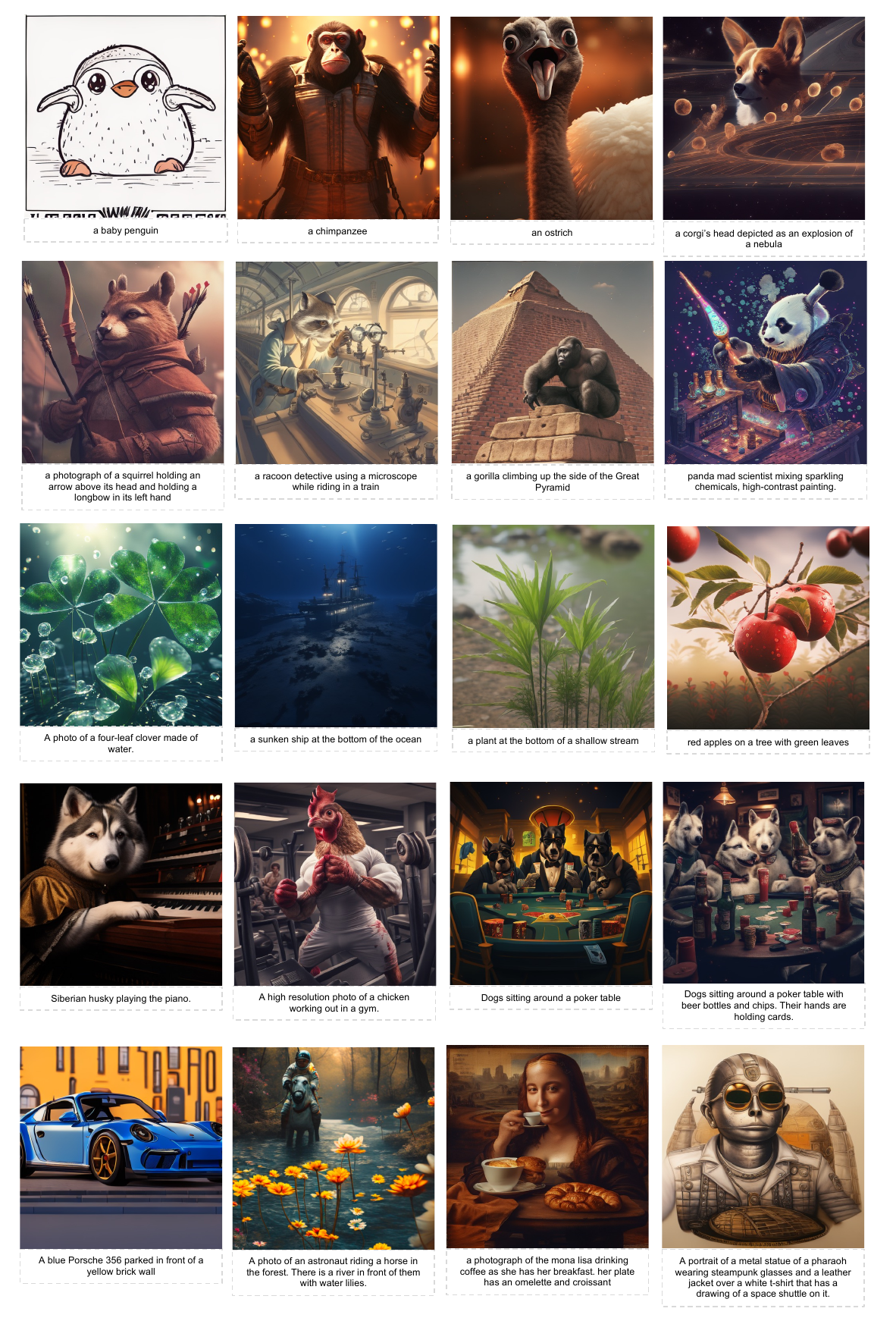}
\caption{(Stage II) Text-conditional 512$\times$512 image generation on PartiPrompts.}
\label{fig:stage2_parti}
\end{figure}
\end{document}